\documentclass[
  onecolumn,
  natbib,
]{miri-tech-article}
\usepackage[
  notodo,
  nocitefix,
]{miritools}
\usepackage[T1]{fontenc}
\usepackage{lmodern}
\usepackage{bm}
\usepackage{enumerate}
\usepackage[noabbrev]{cleveref}
\usepackage[font=small,labelfont=bf]{caption}
\usepackage[hang,flushmargin]{footmisc}
\usepackage{setspace}

\usepackage{tocloft}

\setcitestyle{numbers,square}
\begin{document}

\title{Risks from Learned Optimization \\ in Advanced Machine Learning Systems}
\author{Evan Hubinger\thanks{Alphabetical order. Equal contribution.}, Chris van Merwijk\footnotemark[1], Vladimir Mikulik\footnotemark[1], Joar Skalse\footnotemark[1], and Scott Garrabrant\thanks{With special thanks to Paul Christiano, Eric Drexler, Rob Bensinger, Jan Leike, Rohin Shah, William Saunders, Buck Shlegeris, David Dalrymple, Abram Demski, Stuart Armstrong, Linda Linsefors, Carl Shulman, Toby Ord, Kate Woolverton, and everyone else who provided feedback on earlier versions of this paper.}}

\date{June 11, 2019}


\maketitle

\begin{abstract}
We analyze the type of learned optimization that occurs when a learned model (such as a neural network) is itself an optimizer---a situation we refer to as \textit{mesa-optimization,} a neologism we introduce in this paper. We believe that the possibility of mesa-optimization raises two important questions for the safety and transparency of advanced machine learning systems. First, under what circumstances will learned models be optimizers, including when they should not be? Second, when a learned model is an optimizer, what will its objective be---how will it differ from the loss function it was trained under---and how can it be aligned? In this paper, we provide an in-depth analysis of these two primary questions and provide an overview of topics for future research.
\end{abstract}

\tableofcontents

\section{Introduction}
\label{sec:1}
\cftchapterprecistoc{We introduce the concept of mesa-optimization as well as many relevant terms and concepts related to it, such as what makes a system an optimizer, the difference between base and mesa- optimizers, and our two key problems: unintended optimization and inner alignment.}

In machine learning, we do not manually program each individual parameter of our models. Instead, we specify an objective function that captures what we want the system to do and a learning algorithm to optimize the system for that objective. In this paper, we present a framework that distinguishes what a system is \textit{optimized} to do (its ``purpose''), from what it \textit{optimizes} for (its ``goal''), if it optimizes for anything at all. While all AI systems are optimized for something (have a purpose), whether they actually optimize for anything (pursue a goal) is non-trivial. We will say that a system is an \textit{optimizer} if it is internally searching through a search space (consisting of possible outputs, policies, plans, strategies, or similar) looking for those elements that score high according to some objective function that is explicitly represented within the system. Learning algorithms in machine learning are optimizers because they search through a space of possible parameters---e.g. neural network weights---and improve the parameters with respect to some objective. Planning algorithms are also optimizers, since they search through possible plans, picking those that do well according to some objective.

Whether a system is an optimizer is a property of its internal structure---what algorithm it is physically implementing---and not a property of its input-output behavior. Importantly, the fact that a system's behavior results in some objective being maximized does not make the system an optimizer. For example, a bottle cap causes water to be held inside the bottle, but it is not optimizing for that outcome since it is not running any sort of optimization algorithm.\cite{bottle_caps} Rather, bottle caps have been \textit{optimized} to keep water in place. The optimizer in this situation is the human that designed the bottle cap by searching through the space of possible tools for one to successfully hold water in a bottle. Similarly, image-classifying neural networks are optimized to achieve low error in their classifications, but are not, in general, themselves performing optimization.

However, it is also possible for a neural network to itself run an optimization algorithm. For example, a neural network could run a planning algorithm that predicts the outcomes of potential plans and searches for those it predicts will result in some desired outcome.\footnote{As a concrete example of what a neural network optimizer might look like, consider TreeQN.\cite{treeqn} TreeQN, as described in Farquhar et al., is a Q-learning agent that performs model-based planning (via tree search in a latent representation of the environment states) \textit{as part of its computation of the Q-function.} Though their agent is an optimizer by design, one could imagine a similar algorithm being \textit{learned} by a DQN agent with a sufficiently expressive approximator for the Q function. Universal Planning Networks, as described by Srinivas et al.,\cite{univ_plan_net} provide another example of a learned system that performs optimization, though the optimization there is built-in in the form of SGD via automatic differentiation. However, research such as that in Andrychowicz et al.\cite{grad_by_grad} and Duan et al.\cite{rl2} demonstrate that optimization algorithms can be learned by RNNs, making it possible that a Universal Planning Networks-like agent could be entirely learned---assuming a very expressive model space---including the internal optimization steps. Note that while these examples are taken from reinforcement learning, optimization might in principle take place in any sufficiently expressive learned system.} Such a neural network would itself be an optimizer because it would be searching through the space of possible plans according to some objective function. If such a neural network were produced in training, there would be two optimizers: the learning algorithm that produced the neural network---which we will call the \textit{base optimizer}---and the neural network itself---which we will call the \textit{mesa-optimizer.}\footnote{Previous work in this space has often centered around the concept of ``optimization daemons,''\cite{arbital_daemons} a framework that we believe is potentially misleading and hope to supplant. Notably, the term ``optimization daemon'' came out of discussions regarding the nature of humans and evolution, and, as a result, carries anthropomorphic connotations.}

The possibility of mesa-optimizers has important implications for the safety of advanced machine learning systems. When a base optimizer generates a mesa-optimizer, safety properties of the base optimizer's objective may not transfer to the mesa-optimizer. Thus, we explore two primary questions related to the safety of mesa-optimizers:
\begin{enumerate}
\item \textbf{Mesa-optimization:} Under what circumstances will learned algorithms be optimizers?
\item \textbf{Inner alignment:} When a learned algorithm is an optimizer, what will its objective be, and how can it be aligned?
\end{enumerate}
Once we have introduced our framework in \cref{sec:1}, we will address the first question in \cref{sec:2}, begin addressing the second question in \cref{sec:3}, and finally delve deeper into a specific aspect of the second question in \cref{sec:4}.

\subsection{Base optimizers and mesa-optimizers}
\label{sec:1.1}
\cftchapterprecistoc{What is a mesa-optimizer, and how does it relate to the base optimizer?}

Conventionally, the base optimizer in a machine learning setup is some sort of gradient descent process with the goal of creating a model designed to accomplish some specific task.

Sometimes, this process will also involve some degree of meta-optimization wherein a \textit{meta-optimizer} is tasked with producing a base optimizer that is itself good at optimizing systems to achieve particular goals. Specifically, we will think of a \textit{meta-optimizer} as any system whose task is optimization. For example, we might design a meta-learning system to help tune our gradient descent process.\cite{grad_by_grad} Though the model found by meta-optimization can be thought of as a kind of learned optimizer, it is not the form of learned optimization that we are interested in for this paper. Rather, we are concerned with a different form of learned optimization which we call \textit{mesa-optimization.}

Mesa-optimization is a conceptual dual of meta-optimization---whereas \textit{meta} is Greek for above, \textit{mesa} is Greek for below.\footnote{The word mesa has been proposed as the opposite of meta.\cite{mesa} The duality comes from thinking of meta-optimization as one layer above the base optimizer and mesa-optimization as one layer below.} \textit{Mesa-optimization} occurs when a base optimizer (in searching for algorithms to solve some problem) finds a model that is itself an optimizer, which we will call a \textit{mesa-optimizer.} Unlike meta-optimization, in which the task itself is optimization, mesa-optimization is task-independent, and simply refers to any situation where the internal structure of the model ends up performing optimization because it is instrumentally useful for solving the given task.

In such a case, we will use \textit{base objective} to refer to whatever criterion the base optimizer was using to select between different possible systems and \textit{mesa-objective} to refer to whatever criterion the mesa-optimizer is using to select between different possible outputs. In reinforcement learning (RL), for example, the base objective is generally the expected return. Unlike the base objective, the mesa-objective is not specified directly by the programmers. Rather, the mesa-objective is simply whatever objective was found by the base optimizer that produced good performance on the training environment. Because the mesa-objective is not specified by the programmers, mesa-optimization opens up the possibility of a mismatch between the base and mesa- objectives, wherein the mesa-objective might seem to perform well on the training environment but lead to bad performance off the training environment. We will refer to this case as \textit{pseudo-alignment} below.

There need not always be a mesa-objective since the algorithm found by the base optimizer will not always be performing optimization. Thus, in the general case, we will refer to the model generated by the base optimizer as a \textit{learned algorithm,} which may or may not be a mesa-optimizer.

\begin{figure}[h!]
  \centering
  \includegraphics[width=0.9\textwidth]{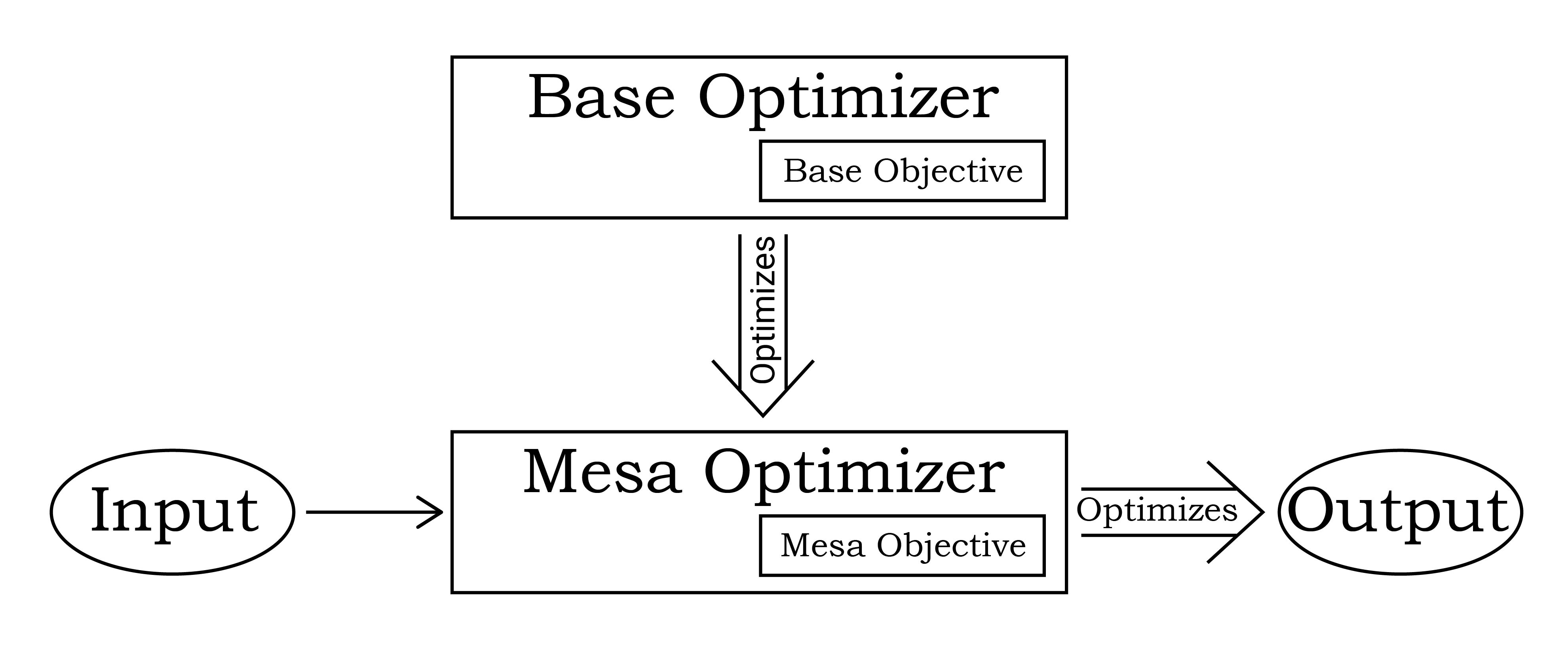}
  \caption{The relationship between the base and mesa- optimizers. The base optimizer optimizes the learned algorithm based on its performance on the base objective. In order to do so, the base optimizer may have turned this learned algorithm into a mesa-optimizer, in which case the mesa-optimizer itself runs an optimization algorithm based on its own mesa-objective. Regardless, it is the learned algorithm that directly takes actions based on its input.}
  \label{fig:1.1}
\end{figure}

\textbf{Possible misunderstanding: ``mesa-optimizer'' does not mean ``subsystem'' or ``subagent.''} In the context of deep learning, a mesa-optimizer is simply a neural network that is implementing some optimization process and not some emergent subagent inside that neural network. Mesa-optimizers are simply a particular type of algorithm that the base optimizer might find to solve its task. Furthermore, we will generally be thinking of the base optimizer as a straightforward optimization algorithm, and not as an intelligent agent choosing to create a subagent.\footnote{That being said, some of our considerations do still apply even in that case.}

We distinguish the mesa-objective from a related notion that we term the \textit{behavioral objective.} Informally, the behavioral objective is the objective which appears to be optimized by the system's behaviour. We can operationalize the behavioral objective as the objective recovered from perfect inverse reinforcement learning (IRL).\footnote{Leike et al.\cite{leike} introduce the concept of an objective recovered from perfect IRL.} This is in contrast to the mesa-objective, which is the objective \textit{actively being used} by the mesa-optimizer in its optimization algorithm.

Arguably, any possible system has a behavioral objective---including bricks and bottle caps. However, for non-optimizers, the appropriate behavioral objective might just be ``1 if the actions taken are those that are in fact taken by the system and 0 otherwise,''\footnote{For the formal construction of this objective, see pg. 6 in Leike et al.\cite{leike}} and it is thus neither interesting nor useful to know that the system is acting to optimize this objective. For example, the behavioral objective ``optimized'' by a bottle cap is the objective of behaving like a bottle cap.\footnote{This objective is by definition trivially optimal in any situation that the bottlecap finds itself in.} However, if the system is an optimizer, then it is more likely that it will have a meaningful behavioral objective. That is, to the degree that a mesa-optimizer's output is systematically selected to optimize its mesa-objective, its behavior may look more like coherent attempts to move the world in a particular direction.\footnote{Ultimately, our worry is optimization in the direction of some coherent but unsafe objective. In this paper, we assume that search provides sufficient structure to expect coherent objectives. While we believe this is a reasonable assumption, it is unclear both whether search is necessary and whether it is sufficient. Further work examining this assumption will likely be needed.}

A given mesa-optimizer's mesa-objective is determined entirely by its internal workings. Once training is finished and a learned algorithm is selected, its direct output---e.g. the actions taken by an RL agent---no longer depends on the base objective. Thus, it is the mesa-objective, not the base objective, that determines a mesa-optimizer's behavioral objective. Of course, to the degree that the learned algorithm was selected on the basis of the base objective, its output will score well on the base objective. However, in the case of a distributional shift, we should expect a mesa-optimizer's behavior to more robustly optimize for the mesa-objective since its behavior is directly computed according to it.

As an example to illustrate the base/mesa distinction in a different domain, and the possibility of misalignment between the base and mesa- objectives, consider biological evolution. To a first approximation, evolution selects organisms according to the objective function of their inclusive genetic fitness in some environment.\footnote{The situation with evolution is more complicated than is presented here and we do not expect our analogy to live up to intense scrutiny. We present it as nothing more than that: an evocative analogy (and, to some extent, an existence proof) that explains the key concepts. More careful arguments are presented later.} Most of these biological organisms---plants, for example---are not ``trying'' to achieve anything, but instead merely implement heuristics that have been pre-selected by evolution. However, some organisms, such as humans, have behavior that does not merely consist of such heuristics but is instead also the result of goal-directed optimization algorithms implemented in the brains of these organisms. Because of this, these organisms can perform behavior that is completely novel from the perspective of the evolutionary process, such as humans building computers.

However, humans tend not to place explicit value on evolution's objective, at least in terms of caring about their alleles' frequency in the population. The objective function stored in the human brain is not the same as the objective function of evolution. Thus, when humans display novel behavior optimized for their own objectives, they can perform very poorly according to evolution's objective. Making a decision not to have children is a possible example of this. Therefore, we can think of evolution as a base optimizer that produced brains---mesa-optimizers---which then actually produce organisms' behavior---behavior that is not necessarily aligned with evolution.

\subsection{The inner and outer alignment problems}
\label{sec:1.2}
\cftchapterprecistoc{We introduce a distinction between the two different, independent alignment problems that appear in the case of mesa-optimization.}

In ``Scalable agent alignment via reward modeling,'' Leike et al. describe the concept of the ``reward-result gap'' as the difference between the (in their case learned) ``reward model'' (what we call the base objective) and the ``reward function that is recovered with perfect inverse reinforcement learning'' (what we call the behavioral objective).\cite{leike} That is, the reward-result gap is the fact that there can be a difference between what a learned algorithm is observed to be doing and what the programmers want it to be doing.

The problem posed by misaligned mesa-optimizers is a kind of reward-result gap. Specifically, it is the gap between the base objective and the mesa-objective (which then causes a gap between the base objective and the behavioral objective). We will call the problem of eliminating the base-mesa objective gap the \textit{inner alignment problem,} which we will contrast with the \textit{outer alignment problem} of eliminating the gap between the base objective and the intended goal of the programmers. This terminology is motivated by the fact that the inner alignment problem is an alignment problem entirely internal to the machine learning system, whereas the outer alignment problem is an alignment problem between the system and the humans outside of it (specifically between the base objective and the programmer's intentions). In the context of machine learning, outer alignment refers to aligning the specified loss function with the intended goal, whereas inner alignment refers to aligning the mesa-objective of a mesa-optimizer with the specified loss function.

It might not be necessary to solve the inner alignment problem in order to produce safe, highly capable AI systems, as it might be possible to prevent mesa-optimizers from occurring in the first place. If mesa-optimizers cannot be reliably prevented, however, then some solution to both the outer and inner alignment problems will be necessary to ensure that mesa-optimizers are aligned with the intended goal of the programmers.

\subsection{Robust alignment vs. pseudo-alignment}
\label{sec:1.3}
\cftchapterprecistoc{Mesa-optimizers need not be robustly aligned with the base optimizer that created them, only pseudo-aligned.}

Given enough training, a mesa-optimizer should eventually be able to produce outputs that score highly on the base objective on the training distribution. Off the training distribution, however---and even on the training distribution while it is still early in the training process---the difference could be arbitrarily large. We will use the term \textit{robustly aligned} to refer to mesa-optimizers with mesa-objectives that robustly agree with the base objective across distributions and the term \textit{pseudo-aligned} to refer to mesa-optimizers with mesa-objectives that agree with the base objective on past training data, but not robustly across possible future data (either in testing, deployment, or further training). For a pseudo-aligned mesa-optimizer, there will be environments in which the base and mesa- objectives diverge. Pseudo-alignment, therefore, presents a potentially dangerous robustness problem since it opens up the possibility of a machine learning system that competently takes actions to achieve something other than the intended goal when off the training distribution. That is, its capabilities might generalize while its objective does not.

For a toy example of what pseudo-alignment might look like, consider an RL agent trained on a maze navigation task where all the doors during training happen to be red. Let the base objective (reward function) be $O_\text{base} = \text{(1 if reached a door, 0 otherwise)}$. On the training distribution, this objective is equivalent to $O_\text{alt} = \text{(1 if reached something red, 0 otherwise)}$. Consider what would happen if an agent, trained to high performance on $O_\text{base}$ on this task, were put in an environment where the doors are instead blue, and with some red objects that are not doors. It might generalize on $O_\text{base}$, reliably navigating to the blue door in each maze (robust alignment). But it might also generalize on $O_\text{alt}$ instead of $O_\text{base}$, reliably navigating each maze to reach red objects (pseudo-alignment).\footnote{Of course, it might also fail to generalize at all.}

\subsection{Mesa-optimization as a safety problem}
\label{sec:1.4}
\cftchapterprecistoc{There are two major safety concerns arising from mesa-optimization: optimization might occur even when it is not intended, and when optimization does occur it might not be optimizing for the right objective.}

If pseudo-aligned mesa-optimizers may arise in advanced ML systems, as we will suggest, they could pose two critical safety problems.

\textbf{Unintended optimization.} First, the possibility of mesa-optimization means that an advanced ML system could end up implementing a powerful optimization procedure even if its programmers never intended it to do so. This could be dangerous if such optimization leads the system to take extremal actions outside the scope of its intended behavior in trying to maximize its mesa-objective. Of particular concern are optimizers with objective functions and optimization procedures that generalize to the real world. The conditions that lead a learning algorithm to find mesa-optimizers, however, are very poorly understood. Knowing them would allow us to predict cases where mesa-optimization is more likely, as well as take measures to discourage mesa-optimization from occurring in the first place. \Cref{sec:2} will examine some features of machine learning algorithms that might influence their likelihood of finding mesa-optimizers.

\textbf{Inner alignment.} Second, even in cases where it is acceptable for a base optimizer to find a mesa-optimizer, a mesa-optimizer might optimize for something other than the specified reward function. In such a case, it could produce bad behavior even if optimizing the correct reward function was known to be safe. This could happen either during training---before the mesa-optimizer gets to the point where it is aligned over the training distribution---or during testing or deployment when the system is off the training distribution. \Cref{sec:3} will address some of the different ways in which a mesa-optimizer could be selected to optimize for something other than the specified reward function, as well as what attributes of an ML system are likely to encourage this. In \cref{sec:4}, we will discuss a possible extreme inner alignment failure---which we believe presents one of the most dangerous risks along these lines---wherein a sufficiently capable misaligned mesa-optimizer could learn to behave as if it were aligned without actually being robustly aligned. We will call this situation \textit{deceptive alignment.}

It may be that pseudo-aligned mesa-optimizers are easy to address---if there exists a reliable method of aligning them, or of preventing base optimizers from finding them. However, it may also be that addressing misaligned mesa-optimizers is very difficult---the problem is not sufficiently well-understood at this point for us to know. Certainly, current ML systems do not produce dangerous mesa-optimizers, though whether future systems might is unknown. It is indeed because of these unknowns that we believe the problem is important to analyze.

\section{Conditions for mesa-optimization}
\label{sec:2}
\cftchapterprecistoc{Not all machine learning systems exhibit mesa-optimization, but some might. We seek to understand what sorts of machine learning systems are likely to exhibit mesa-optimization and what sorts are not.}

In this section, we consider how the following two components of a particular machine learning system might influence whether it will produce a mesa-optimizer:
\begin{enumerate}
\item \textbf{The task:} The training distribution and base objective function.
\item \textbf{The base optimizer:} The machine learning algorithm and model architecture.
\end{enumerate}

We deliberately choose to present theoretical considerations for why mesa-optimization may or may not occur rather than provide concrete examples. Mesa-optimization is a phenomenon that we believe will occur mainly in machine learning systems that are more advanced than those that exist today.\footnote{As of the date of this paper. Note that we do examine some existing machine learning systems that we believe are close to producing mesa-optimization in \cref{sec:5}.} Thus, an attempt to induce mesa-optimization in a \textit{current} machine learning system would likely require us to use an artificial setup specifically designed to induce mesa-optimization. Moreover, the limited interpretability of neural networks, combined with the fact that there is no general and precise definition of ``optimizer,'' means that it would be hard to evaluate whether a given model is a mesa-optimizer.

\subsection{The task}
\label{sec:2.1}
\cftchapterprecistoc{How does the task being solved contribute to the likelihood of a machine learning system exhibiting mesa-optimization?}

Some tasks benefit from mesa-optimizers more than others. For example, tic-tac-toe can be perfectly solved by simple rules. Thus, a base optimizer has no need to generate a mesa-optimizer to solve tic-tac-toe, since a simple learned algorithm implementing the rules for perfect play will do. Human survival in the savanna, by contrast, did seem to benefit from mesa-optimization. Below, we discuss the properties of tasks that may influence the likelihood of mesa-optimization.

\textbf{Better generalization through search.} To be able to consistently achieve a certain level of performance in an environment, we hypothesize that there will always have to be some minimum amount of optimization power that must be applied to find a policy that performs that well.

To see this, we can think of optimization power as being measured in terms of the number of times the optimizer is able to divide the search space in half---that is, the number of bits of information provided.\cite{optpow} After these divisions, there will be some remaining space of policies that the optimizer is unable to distinguish between. Then, to ensure that all policies in the remaining space have some minimum level of performance---to provide a performance lower bound\footnote{It is worth noting that the same argument also holds for achieving an average-case guarantee.}---will always require the original space to be divided some minimum number of times---that is, there will always have to be some minimum bits of optimization power applied.

However, there are two distinct levels at which this optimization power could be expended: the base optimizer could expend optimization power selecting a highly-tuned learned algorithm, or the learned algorithm could itself expend optimization power selecting highly-tuned actions.

As a mesa-optimizer is just a learned algorithm that itself performs optimization, the degree to which mesa-optimizers will be incentivized in machine learning systems is likely to be dependent on which of these levels it is more advantageous for the system to perform optimization. For many current machine learning models, where we expend vastly more computational resources training the model than running it, it seems generally favorable for most of the optimization work to be done by the base optimizer, with the resulting learned algorithm being simply a network of highly-tuned heuristics rather than a mesa-optimizer.

We are already encountering some problems, however---Go, Chess, and Shogi, for example---for which this approach does not scale. Indeed, our best current algorithms for those tasks involve explicitly making an optimizer (hard-coded Monte-Carlo tree search with learned heuristics) that does optimization work on the level of the learned algorithm rather than having all the optimization work done by the base optimizer.\cite{alphazero} Arguably, this sort of task is only adequately solvable this way---if it were possible to train a straightforward DQN agent to perform well at Chess, it plausibly would \textit{have} to learn to internally perform something like a tree search, producing a mesa-optimizer.\footnote{Assuming reasonable computational constraints.}

We hypothesize that the attractiveness of search in these domains is due to the diverse, branching nature of these environments. This is because search---that is, optimization---tends to be good at generalizing across diverse environments, as it gets to individually determine the best action for each individual task instance. There is a general distinction along these lines between optimization work done on the level of the learned algorithm and that done on the level of the base optimizer: the learned algorithm only has to determine the best action for a given task instance, whereas the base optimizer has to design heuristics that will hold regardless of what task instance the learned algorithm encounters. Furthermore, a mesa-optimizer can immediately optimize its actions in novel situations, whereas the base optimizer can only change the mesa-optimizer's policy by modifying it ex-post. Thus, for environments that are diverse enough that most task instances are likely to be completely novel, search allows the mesa-optimizer to adjust for that new task instance immediately.

For example, consider reinforcement learning in a diverse environment, such as one that directly involves interacting with the real world. We can think of a diverse environment as requiring a very large amount of computation to figure out good policies before conditioning on the specifics of an individual instance, but only a much smaller amount of computation to figure out a good policy once the specific instance of the environment is known. We can model this observation as follows.

Suppose an environment is composed of $N$ different instances, each of which requires a completely distinct policy to succeed in.\footnote{This definition of $N$ is somewhat vague, as there are multiple different levels at which one can chunk an environment into instances. For example, one environment could always have the same high-level features but completely random low-level features, whereas another could have two different categories of instances that are broadly self-similar but different from each other, in which case it's unclear which has a larger $N$. However, one can simply imagine holding $N$ constant for all levels but one and just considering how environment diversity changes on that level.} Let $P$ be the optimization power (measured in bits\cite{optpow}) applied by the base optimizer, which should be approximately proportional to the number of training steps. Then, let $x$ be the optimization power applied by the learned algorithm in each environment instance and $f(x)$ the total amount of optimization power the base optimizer must put in to get a learned algorithm capable of performing that amount of optimization.\footnote{Note that this makes the implicit assumption that the amount of optimization power required to find a mesa-optimizer capable of performing $x$ bits of optimization is independent of $N$. The justification for this is that optimization is a general algorithm that looks the same regardless of what environment it is applied to, so the amount of optimization required to find an $x$-bit optimizer should be relatively independent of the environment. That being said, it won't be completely independent, but as long as the primary difference between environments is how much optimization they need, rather than how hard it is to do optimization, the model presented here should hold.} We will assume that the rest of the base optimizer's optimization power, $P - f(x)$, goes into tuning the learned algorithm's policy. Since the base optimizer has to distribute its tuning across all $N$ task instances, the amount of optimization power it will be able to contribute to each instance will be $\frac{P - f(x)}{N}$, under the previous assumption that each instance requires a completely distinct policy. On the other hand, since the learned algorithm does all of its optimization at runtime, it can direct all of it into the given task instance, making its contribution to the total for each instance simply $x$.\footnote{Note, however, that there will be some maximum $x$ simply because the learned algorithm generally only has access to so much computational power.}

Thus, if we assume that, for a given $P$, the base optimizer will select the value of $x$ that maximizes the minimum level of performance, and thus the total optimization power applied to each instance, we get\footnote{Subject to the constraint that $P - f(x) \ge 0$.}
\[
  x^* = \argmax_x~ \frac{P - f(x)}{N} + x.
\]
As one moves to more and more diverse environments---that is, as $N$ increases---this model suggests that $x$ will dominate $\frac{P - f(x)}{N}$, implying that mesa-optimization will become more and more favorable. Of course, this is simply a toy model, as it makes many questionable simplifying assumptions. Nevertheless, it sketches an argument for a pull towards mesa-optimization in sufficiently diverse environments.

As an illustrative example, consider biological evolution. The environment of the real world is highly diverse, resulting in non-optimizer policies directly fine-tuned by evolution---those of plants, for example---having to be very simple, as evolution has to spread its optimization power across a very wide range of possible environment instances. On the other hand, animals with nervous systems can display significantly more complex policies by virtue of being able to perform their own optimization, which can be based on immediate information from their environment. This allows sufficiently advanced mesa-optimizers, such as humans, to massively outperform other species, especially in the face of novel environments, as the optimization performed internally by humans allows them to find good policies even in entirely novel environments.

\textbf{Compression of complex policies.} In some tasks, good performance requires a very complex policy. At the same time, base optimizers are generally biased in favor of selecting learned algorithms with lower complexity. Thus, all else being equal, the base optimizer will generally be incentivized to look for a highly compressed policy.

One way to find a compressed policy is to search for one that is able to use general features of the task structure to produce good behavior, rather than simply memorizing the correct output for each input. A mesa-optimizer is an example of such a policy. From the perspective of the base optimizer, a mesa-optimizer is a highly-compressed version of whatever policy it ends up implementing: instead of explicitly encoding the details of that policy in the learned algorithm, the base optimizer simply needs to encode how to search for such a policy. Furthermore, if a mesa-optimizer can determine the important features of its environment at runtime, it does not need to be given as much prior information as to what those important features are, and can thus be much simpler.

This effect is most pronounced for tasks with a broad diversity of details but common high-level features. For example, Go, Chess, and Shogi have a very large domain of possible board states, but admit a single high-level strategy for play---heuristic-guided tree search---that performs well across all board states.\cite{alphazero} On the other hand, a classifier trained on random noise is unlikely to benefit from compression at all.

The environment need not necessarily be too diverse for this sort of effect to appear, however, as long as the pressure for low description length is strong enough. As a simple illustrative example, consider the following task: given a maze, the learned algorithm must output a path through the maze from start to finish. If the maze is sufficiently long and complicated then the specific strategy for solving this particular maze---specifying each individual turn---will have a high description length. However, the description length of a general optimization algorithm for finding a path through an arbitrary maze is fairly small. Therefore, if the base optimizer is selecting for programs with low description length, then it might find a mesa-optimizer that can solve all mazes, despite the training environment only containing one maze.

\textbf{Task restriction.} The observation that diverse environments seem to increase the probability of mesa-optimization suggests that one way of reducing the probability of mesa-optimizers might be to keep the tasks on which AI systems are trained highly restricted. Focusing on building many individual AI services which can together offer all the capabilities of a generally-intelligent system rather than a single general-purpose artificial general intelligence (AGI), for example, might be a way to accomplish this while still remaining competitive with other approaches.\cite{drexler}

\textbf{Human modeling.} Another aspect of the task that might influence the likelihood of mesa-optimization is the presence of humans in the environment.\cite{human_models} Since humans often act as optimizers, reasoning about humans will likely involve reasoning about optimization. A system capable of reasoning about optimization is likely also capable of reusing that same machinery to do optimization itself, resulting in a mesa-optimizer. For example, it might be the case that predicting human behavior requires instantiating a process similar to human judgment, complete with internal motives for making one decision over another.

Thus, tasks that do not benefit from human modeling seem less likely to produce mesa-optimizers than those that do. Furthermore, there are many tasks that might benefit from human modeling that don't explicitly involve modeling humans---to the extent that the training distribution is generated by humans, for example, modeling humans might enable the generation of a very good prior for that distribution.

\subsection{The base optimizer}
\label{sec:2.2}
\cftchapterprecistoc{How does the machine learning algorithm and model architecture contribute to the likelihood of a machine learning system exhibiting mesa-optimization?}

It is likely that certain features of the base optimizer will influence how likely it is to generate a mesa-optimizer. First, though we largely focus on reinforcement learning in this paper, RL is not necessarily the only type of machine learning where mesa-optimizers could appear. For example, it seems plausible that mesa-optimizers could appear in generative adversarial networks.

Second, we hypothesize that the details of a machine learning model's architecture will have a significant effect on its tendency to implement mesa-optimization. For example, a tabular model, which independently learns the correct output for every input, will never be a mesa-optimizer. On the other hand, if a hypothetical base optimizer looks for the program with the shortest source code that solves a task, then it is more plausible that this program will itself be an optimizer.\cite{paul_solomonoff} However, for realistic machine learning base optimizers, it is less clear to what extent mesa-optimizers will be selected for. Thus, we discuss some factors below that might influence the likelihood of mesa-optimization one way or the other.

\textbf{Reachability.} There are many kinds of optimization algorithms that a base optimizer could implement. However, almost every training strategy currently used in machine learning uses some form of local search (such as gradient descent or even genetic algorithms). Thus, it seems plausible that the training strategy of more advanced ML systems will also fall into this category. We will call this general class of optimizers that are based on local hill-climbing \textit{local optimization processes.}

We can then formulate a notion of \textit{reachability,} the difficulty for the base optimizer to find any given learned algorithm, which we can analyze in the case of a local optimization process. A local optimization process might fail to find a particular learned algorithm that would perform very well on the base objective if the learned algorithm is surrounded by other algorithms that perform poorly on the base objective. For a mesa-optimizer to be produced by a local optimization process, it needs to not only perform well on the base objective, but also be \textit{reachable;} that is, there needs to be a path through the space of learned algorithms to it that is approximately monotonically increasing. Furthermore, the degree to which the path only need be approximate---that is, the degree to which ML training procedures can escape local optima---is likely to be critical, as optimization algorithms are complex enough that it might require a significant portion of the algorithm to be present before performance gains start being realized.

\textbf{Model capacity.} One key factor likely to determine the reachability of mesa-optimizers in a given machine learning system is its model capacity---that is, how extensive is the set of algorithms (or how expressive is the model space) capable of being found by the base optimizer. The larger the model capacity, the broader the space of possible learned algorithms, and thus the more likely that it will be able to find one that is a mesa-optimizer, assuming the base optimizer is incentivized to do so. For example, architectures that explicitly give the algorithm access to a wide range of possible computations, such as recurrent neural networks or neural Turing machines,\cite{neural_tms} seem more likely to produce mesa-optimizers. Furthermore, the size of a model imposes a hard limit on the complexity of the functions it is able to represent. Thus, to the degree that the base optimizer is selecting based on performance, it will be driven to ``squeeze out'' as much performance as it can for any given model capacity, leading to an inductive bias in favor of relatively compressed policies.

\textbf{Inductive biases.} Another important factor is the degree to which the base optimizer is explicitly or implicitly biased in various ways. The nature of these inductive biases will contribute to the likelihood of a mesa-optimizer being selected for. One of the most important kinds of inductive bias is simplicity bias, which would almost certainly be exhibited by almost all base optimizers. We identify two basic ways in which simplicity bias can manifest itself, in addition to the one due to model capacity as mentioned above:
\begin{enumerate}
\item An explicit penalty due to parameter regularization or architectural constraints such as weight-sharing or sparse connections.
\item An implicit bias due to the model architecture. For example, it has been shown that neural networks are more likely to fit a simple function to a set of training data, even when no regularization is used.\cite{nn_simp_bias}
\end{enumerate}
The more a base optimizer is biased towards simple solutions, the more it will be incentivized to find a compressed policy such as a mesa-optimizer.

The base optimizer could also be biased in other ways. For example, it could be biased towards algorithms with a low time or space complexity. If we view mesa-optimization as delegation of optimization work, a time penalty on the learned algorithm is likely to incentivize the base optimizer to do more precomputation itself by pre-computing relevant heuristics rather than delegating optimization work to the learned algorithm. Thus, we hypothesize that penalizing description length will favor mesa-optimizers while penalizing time complexity will disfavor them. This suggests the following speculative conjecture: neither a minimal-depth nor minimal-size boolean circuit that solves a problem can be a mesa-optimizer.\cite{paul_minimal_circuits}

Lastly, another form of bias that might have particularly interesting effects is the presence of an information funnel. In deep learning, the base optimizer directly selects the values of individual parameters. In biological evolution, by contrast, the base optimizer selects DNA strings, which in turn produce the detailed structure of the brain only indirectly. This implicitly introduces pressure towards compressing the brain's structure. As we noted previously, this might favor the development of mesa-optimizers, as they can be seen as a way of compressing complex policies. Because deep learning algorithms currently select weights directly, there is no such information funnel. In the case of meta-learning over a machine learning model's architecture, however, an information funnel might present itself due to the fact that the meta-learner has to learn a single policy for creating architectures that is capable of solving all of the problems it is presented with.

\textbf{Statefulness.} Another factor that could make it easier for a learned algorithm to implement a complex optimization process is the ability of the learned algorithm to save and recall information. The ability to refer back to previous state enables the caching of intermediate results, increasing the model capacity by giving the learned algorithm significantly more time to complete its computations. Statefulness, therefore, favors mesa-optimization by decreasing the implicit penalty on time complexity imposed by enforcing all computations to end when the learned algorithm produces its output. One simple form of this might be a recurrent neural network, which can perform computations with longer time horizons due to its ability to pass intermediate activations across different time steps. An even more extreme version of this might be a network that is able to store state not just across different time steps in the same episode, but also across episodes, enabling it to do more long-term learning, though such an architecture is currently completely hypothetical.

\textbf{Hard-coded optimization.} One possible means of alleviating some of these issues might be to include hard-coded optimization where the learned algorithm provides only the objective function and not the optimization algorithm. The stronger the optimization performed explicitly, the less strong the optimization performed implicitly by the learned algorithm needs to be. For example, architectures that explicitly perform optimization that is relevant for the task---such as hard-coded Monte Carlo tree search---might decrease the benefit of mesa-optimizers by reducing the need for optimization other than that which is explicitly programmed into the system.

\section{The inner alignment problem}
\label{sec:3}
\cftchapterprecistoc{We detail some of the ways in which a mesa-optimizer can be misaligned with the base objective and analyze factors that might influence the probability of misalignment.}

In this section, we outline reasons to think that a mesa-optimizer may not optimize the same objective function as its base optimizer. Machine learning practitioners have direct control over the base objective function---either by specifying the loss function directly or training a model for it---but cannot directly specify the mesa-objective developed by a mesa-optimizer. We refer to this problem of aligning mesa-optimizers with the base objective as the inner alignment problem. This is distinct from the outer alignment problem, which is the traditional problem of ensuring that the base objective captures the intended goal of the programmers.

Current machine learning methods select learned algorithms by empirically evaluating their performance on a set of training data according to the base objective function. Thus, ML base optimizers select mesa-optimizers according to the output they produce rather than directly selecting for a particular mesa-objective. Moreover, the selected mesa-optimizer's policy only has to perform well (as scored by the base objective) on the training data. If we adopt the assumption that the mesa-optimizer computes an optimal policy given its objective function, then we can summarize the relationship between the base and mesa- objectives as follows:\cite{chris}
\begin{align*}
  \theta^* &= \argmax_\theta~ \mathbb E(O_{\text{base}}(\pi_\theta)),~ \text{where} \\
  \pi_\theta &= \argmax_\pi~ \mathbb E(O_{\text{mesa}}(\pi \,|\, \theta))
\end{align*}
That is, the base optimizer maximizes its objective $O_\text{base}$ by choosing a mesa-optimizer with parameterization $\theta$ based on the mesa-optimizer's policy $\pi_\theta$, but not based on the objective function $O_\text{mesa}$ that the mesa-optimizer uses to compute this policy. Depending on the base optimizer, we will think of $O_\text{base}$ as the negative of the loss, the future discounted reward, or simply some fitness function by which learned algorithms are being selected.

An interesting approach to analyzing this connection is presented in Ibarz et al, where empirical samples of the true reward and a learned reward on the same trajectories are used to create a scatter-plot visualization of the alignment between the two.\cite{ibarz} The assumption in that work is that a monotonic relationship between the learned reward and true reward indicates alignment, whereas deviations from that suggest misalignment. Building on this sort of research, better theoretical measures of alignment might someday allow us to speak concretely in terms of provable guarantees about the extent to which a mesa-optimizer is aligned with the base optimizer that created it.

\subsection{Pseudo-alignment}
\label{sec:3.1}
\cftchapterprecistoc{What are the different ways in which a mesa-optimizer can be pseudo-aligned?}

There is currently no complete theory of the factors that affect whether a mesa-optimizer will be pseudo-aligned---that is, whether it will appear aligned on the training data, while actually optimizing for something other than the base objective. Nevertheless, we outline a basic classification of ways in which a mesa-optimizer could be pseudo-aligned:
\begin{enumerate}
\item \textbf{Proxy alignment,}
\item \textbf{Approximate alignment,} and
\item \textbf{Suboptimality alignment.}
\end{enumerate}

\textbf{Proxy alignment.} The basic idea of \textit{proxy alignment} is that a mesa-optimizer can learn to optimize for some proxy of the base objective instead of the base objective itself. We'll start by considering two special cases of proxy alignment: \textit{side-effect alignment} and \textit{instrumental alignment.}

First, a mesa-optimizer is \textit{side-effect aligned} if optimizing for the mesa-objective $O_\text{mesa}$ has the direct causal result of increasing the base objective $O_\text{base}$ in the training distribution, and thus when the mesa-optimizer optimizes $O_\text{mesa}$ it results in an increase in $O_\text{base}$. For an example of side-effect alignment, suppose that we are training a cleaning robot. Consider a robot that optimizes the number of times it has swept a dusty floor. Sweeping a floor causes the floor to be cleaned, so this robot would be given a good score by the base optimizer. However, if during deployment it is offered a way to make the floor dusty again after cleaning it (e.g. by scattering the dust it swept up back onto the floor), the robot will take it, as it can then continue sweeping dusty floors.

Second, a mesa-optimizer is \textit{instrumentally aligned} if optimizing for the base objective $O_\text{base}$ has the direct causal result of increasing the mesa-objective $O_\text{mesa}$ in the training distribution, and thus the mesa-optimizer optimizes $O_\text{base}$ as an instrumental goal for the purpose of increasing $O_\text{mesa}$. For an example of instrumental alignment, suppose again that we are training a cleaning robot. Consider a robot that optimizes the amount of dust in the vacuum cleaner. Suppose that in the training distribution the easiest way to get dust into the vacuum cleaner is to vacuum the dust on the floor. It would then do a good job of cleaning in the training distribution and would be given a good score by the base optimizer. However, if during deployment the robot came across a more effective way to acquire dust---such as by vacuuming the soil in a potted plant---then it would no longer exhibit the desired behavior.

We propose that it is possible to understand the general interaction between side-effect and instrumental alignment using causal graphs, which leads to our general notion of proxy alignment.

Suppose we model a task as a causal graph with nodes for all possible attributes of that task and arrows between nodes for all possible relationships between those attributes. Then we can also think of the mesa-objective $O_\text{mesa}$ and the base objective $O_\text{base}$ as nodes in this graph. For $O_\text{mesa}$ to be pseudo-aligned, there must exist some node $X$ such that $X$ is an ancestor of both $O_\text{mesa}$ and $O_\text{base}$ in the training distribution, and such that $O_\text{mesa}$ and $O_\text{base}$ increase with $X$. If $X = O_\text{mesa}$, this is side-effect alignment, and if $X = O_\text{base}$, this is instrumental alignment.

\begin{figure}[h!]
  \centering
  \includegraphics[width=0.5\textwidth]{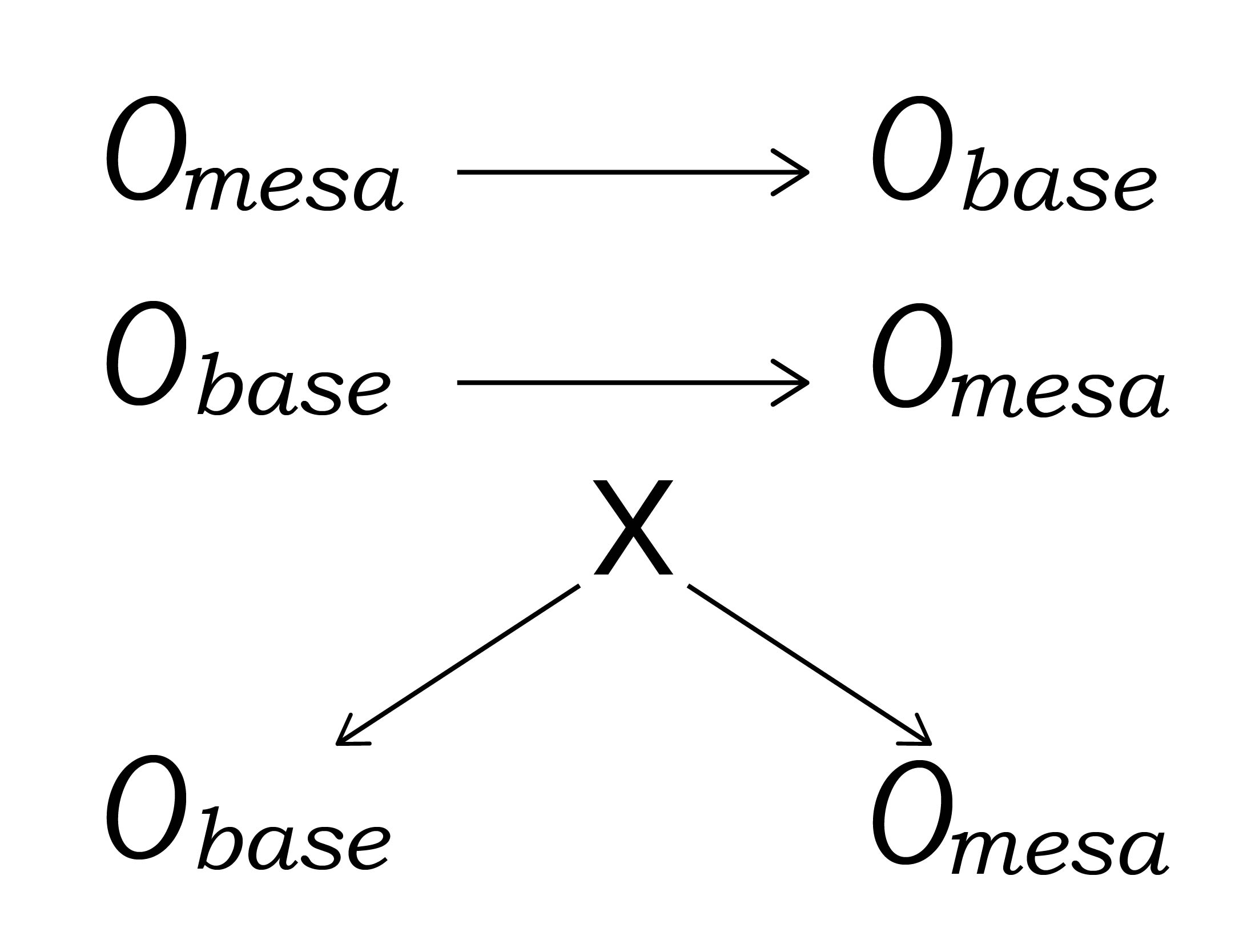}
  \caption{A causal diagram of the training environment for the different types of proxy alignment. The diagrams represent, from top to bottom, side-effect alignment (top), instrumental alignment (middle), and general proxy alignment (bottom). The arrows represent positive causal relationships---that is, cases where an increase in the parent causes an increase in the child.}
  \label{fig:3.1}
\end{figure}

This represents the most generalized form of a relationship between $O_\text{mesa}$ and $O_\text{base}$ that can contribute to pseudo-alignment. Specifically, consider the causal graph given in \cref{fig:3.1}. A mesa-optimizer with mesa-objective $O_\text{mesa}$ will decide to optimize $X$ as an instrumental goal of optimizing $O_\text{mesa}$, since $X$ increases $O_\text{mesa}$. This will then result in $O_\text{base}$ increasing, since optimizing for $X$ has the side-effect of increasing $O_\text{base}$. Thus, in the general case, side-effect and instrumental alignment can work together to contribute to pseudo-alignment over the training distribution, which is the general case of proxy alignment.

\textbf{Approximate alignment.} A mesa-optimizer is \textit{approximately aligned} if the mesa-objective $O_\text{mesa}$ and the base objective $O_\text{base}$ are approximately the same function up to some degree of approximation error related to the fact that the mesa-objective has to be represented inside the mesa-optimizer rather than being directly programmed by humans. For example, suppose you task a neural network with optimizing for some base objective that is impossible to perfectly represent in the neural network itself. Even if you get a mesa-optimizer that is as aligned as possible, it still will not be perfectly robustly aligned in this scenario, since there will have to be some degree of approximation error between its internal representation of the base objective and the actual base objective.

\textbf{Suboptimality alignment.} A mesa-optimizer is \textit{suboptimality aligned} if some deficiency, error, or limitation in its optimization process causes it to exhibit aligned behavior on the training distribution. This could be due to computational constraints, unsound reasoning, a lack of information, irrational decision procedures, or any other defect in the mesa-optimizer's reasoning process. Importantly, we are not referring to a situation where the mesa-optimizer is robustly aligned but nonetheless makes mistakes leading to bad outcomes on the base objective. Rather, suboptimality alignment refers to the situation where the mesa-optimizer is misaligned but nevertheless performs \textit{well} on the base objective, precisely because it has been selected to make mistakes that lead to good outcomes on the base objective.

For an example of suboptimality alignment, consider a cleaning robot with a mesa-objective of minimizing the total amount of stuff in existence. If this robot has the mistaken belief that the dirt it cleans is completely destroyed, then it may be useful for cleaning the room despite doing so not actually helping it succeed at its objective. This robot will be observed to be a good optimizer of $O_\text{base}$ and hence be given a good score by the base optimizer. However, if during deployment the robot is able to improve its world model, it will stop exhibiting the desired behavior.

As another, perhaps more realistic example of suboptimality alignment, consider a mesa-optimizer with a mesa-objective $O_\text{mesa}$ and an environment in which there is one simple strategy and one complicated strategy for achieving $O_\text{mesa}$. It could be that the simple strategy is aligned with the base optimizer, but the complicated strategy is not. The mesa-optimizer might then initially only be aware of the simple strategy, and thus be suboptimality aligned, until it has been run for long enough to come up with the complicated strategy, at which point it stops exhibiting the desired behavior.

\subsection{The task}
\label{sec:3.2}
\cftchapterprecistoc{How does the task being solved contribute to the likelihood of mesa-optimizers being pseudo-aligned?}

As in \cref{sec:2}, we will now consider the task the machine learning system is trained on. Specifically, we will address how the task affects a machine learning system's propensity to produce \textit{pseudo-aligned} mesa-optimizers.

\textbf{Unidentifiability.} It is a common problem in machine learning for a dataset to not contain enough information to adequately pinpoint a specific concept. This is closely analogous to the reason that machine learning models can fail to generalize or be susceptible to adversarial examples\cite{adversarial_examples}---there are many more ways of classifying data that do well in training than any specific way the programmers had in mind. In the context of mesa-optimization, this manifests as pseudo-alignment being more likely to occur when a training environment does not contain enough information to distinguish between a wide variety of different objective functions. In such a case there will be many more ways for a mesa-optimizer to be pseudo-aligned than robustly aligned---one for each indistinguishable objective function. Thus, most mesa-optimizers that do well on the base objective will be pseudo-aligned rather than robustly aligned. This is a critical concern because it makes every other problem of pseudo-alignment worse---it is a reason that, in general, it is hard to find robustly aligned mesa-optimizers. Unidentifiability in mesa-optimization is partially analogous to the problem of unidentifiability in reward learning, in that the central issue is identifying the ``correct'' objective function given particular training data.\cite{irl_unidentifiability} We discuss this relationship further in \cref{sec:5}.

In the context of mesa-optimization, there is also an additional source of unidentifiability stemming from the fact that the mesa-optimizer is selected merely on the basis of its output. Consider the following toy reinforcement learning example. Suppose that in the training environment, pressing a button always causes a lamp to turn on with a ten-second delay, and that there is no other way to turn on the lamp. If the base objective depends only on whether the lamp is turned on, then a mesa-optimizer that maximizes button presses and one that maximizes lamp light will show identical behavior, as they will both press the button as often as they can. Thus, we cannot distinguish these two objective functions in this training environment. Nevertheless, the training environment does contain enough information to distinguish at least between these two particular objectives: since the high reward only comes after the ten-second delay, it must be from the lamp, not the button. As such, even if a training environment in principle contains enough information to identify the base objective, it might still be impossible to distinguish robustly aligned from proxy-aligned mesa-optimizers.

\textbf{Proxy choice as pre-computation.} Proxy alignment can be seen as a form of pre-computation by the base optimizer. Proxy alignment allows the base optimizer to save the mesa-optimizer computational work by pre-computing which proxies are valuable for the base objective and then letting the mesa-optimizer maximize those proxies.

Without such pre-computation, the mesa-optimizer has to infer at runtime the causal relationship between different input features and the base objective, which might require significant computational work. Moreover, errors in this inference could result in outputs that perform worse on the base objective than if the system had access to pre-computed proxies. If the base optimizer precomputes some of these causal relationships---by selecting the mesa-objective to include good proxies---more computation at runtime can be diverted to making better plans instead of inferring these relationships.

The case of biological evolution may illustrate this point. The proxies that humans care about---food, resources, community, mating, etc.---are relatively computationally easy to optimize directly, while correlating well with survival and reproduction in our ancestral environment. For a human to be robustly aligned with evolution would have required us to instead care directly about spreading our genes, in which case we would have to infer that eating, cooperating with others, preventing physical pain, etc. would promote genetic fitness in the long run, which is not a trivial task. To infer all of those proxies from the information available to early humans would have required greater (perhaps unfeasibly greater) computational resources than to simply optimize for them directly. As an extreme illustration, for a child in this alternate universe to figure out not to stub its toe, it would have to realize that doing so would slightly diminish its chances of reproducing twenty years later.

For pre-computation to be beneficial, there needs to be a relatively stable causal relationship between a proxy variable and the base objective such that optimizing for the proxy will consistently do well on the base objective. However, even an imperfect relationship might give a significant performance boost over robust alignment if it frees up the mesa-optimizer to put significantly more computational effort into optimizing its output. This analysis suggests that there might be pressure towards proxy alignment in complex training environments, since the more complex the environment, the more computational work pre-computation saves the mesa-optimizer. Additionally, the more complex the environment, the more potential proxy variables are available for the mesa-optimizer to use.

Furthermore, in the context of machine learning, this analysis suggests that a time complexity penalty (as opposed to a description length penalty) is a double-edged sword. In \cref{sec:2}, we suggested that penalizing time complexity might serve to reduce the likelihood of mesa-optimization. However, the above suggests that doing so would also promote pseudo-alignment in those cases where mesa-optimizers do arise. If the cost of fully modeling the base objective in the mesa-optimizer is large, then a pseudo-aligned mesa-optimizer might be preferred simply because it reduces time complexity, even if it would underperform a robustly aligned mesa-optimizer without such a penalty.

\textbf{Compression of the mesa-optimizer.} The description length of a robustly aligned mesa-optimizer may be greater than that of a pseudo-aligned mesa-optimizer. Since there are more pseudo-aligned mesa-objectives than robustly aligned mesa-objectives, pseudo-alignment provides more degrees of freedom for choosing a particularly simple mesa-objective. Thus, we expect that in most cases there will be several pseudo-aligned mesa-optimizers that are less complex than any robustly aligned mesa-optimizer.

This description cost is especially high if the learned algorithm's input data does not contain easy-to-infer information about how to optimize for the base objective. Biological evolution seems to differ from machine learning in this sense, since evolution's specification of the brain has to go through the information funnel of DNA. The sensory data that early humans received didn't allow them to infer the existence of DNA, nor the relationship between their actions and their genetic fitness. Therefore, for humans to have been aligned with evolution would have required them to have an innately specified model of DNA, as well as the various factors influencing their inclusive genetic fitness. Such a model would not have been able to make use of environmental information for compression, and thus would have required a greater description length. In contrast, our models of food, pain, etc. can be very short since they are directly related to our input data.

\subsection{The base optimizer}
\label{sec:3.3}
\cftchapterprecistoc{How does the machine learning algorithm and model architecture contribute to the likelihood of mesa-optimizers being pseudo-aligned?}

We now turn to how the base optimizer is likely to affect the propensity for a machine learning system to produce pseudo-aligned mesa-optimizers.

\textbf{Hard-coded optimization.} In \cref{sec:2}, we suggested that hard-coding an optimization algorithm---that is to say, choosing a model with built-in optimization---could be used to remove some of the incentives for mesa-optimization. Similarly, hard-coded optimization may be used to prevent some of the sources of pseudo-alignment, since it may allow one to directly specify or train the mesa-objective. Reward-predictive model-based reinforcement learning might be one possible way of accomplishing this.\cite{imagination_planners} For example, an ML system could include a model directly trained to predict the base objective together with a powerful hard-coded optimization algorithm. Doing this bypasses some of the problems of pseudo-alignment: if the mesa-optimizer is trained to directly predict the base reward, then it will be selected to make good predictions even if a bad prediction would result in a good policy. However, a learned model of the base objective will still be underdetermined off-distribution, so this approach by itself does not guarantee robust alignment.

\textbf{Model capacity.} We hypothesize that a system's model capacity will have implications for how likely it is to develop pseudo-aligned learned algorithms. One possible source of pseudo-alignment that could be particularly difficult to avoid is approximation error---if a mesa-optimizer is not capable of faithfully representing the base objective, then it can't possibly be robustly aligned, only approximately aligned. Even if a mesa-optimizer might theoretically be able to perfectly capture the base objective, the more difficult that is for it to do, the more we might expect it to be approximately aligned rather than robustly aligned. Thus, a large model capacity may be both a blessing and a curse: it makes it less likely that mesa-optimizers will be approximately aligned, but it also increases the likelihood of getting a mesa-optimizer in the first place.\footnote{Though a large model capacity seems to make approximate alignment less likely, it is unclear how it might affect other forms of pseudo-alignment such as deceptive alignment.}

\textbf{Subprocess interdependence.} There are some reasons to believe that there might be more initial optimization pressure towards proxy aligned than robustly aligned mesa-optimizers. In a local optimization process, each parameter of the learned algorithm (e.g. the parameter vector of a neuron) is adjusted to locally improve the base objective \textit{conditional} on the other parameters. Thus, the benefit for the base optimizer of developing a new subprocess will likely depend on what other subprocesses the learned algorithm currently implements. Therefore, even if some subprocess would be very beneficial if combined with many other subprocesses, the base optimizer may not select for it until the subprocesses it depends on are sufficiently developed. As a result, a local optimization process would likely result in subprocesses that have fewer dependencies being developed before those with more dependencies.

In the context of mesa-optimization, the benefit of a robustly aligned mesa-objective seems to depend on more subprocesses than at least some pseudo-aligned mesa-objectives. For example, consider a side-effect aligned mesa-optimizer optimizing for some set of proxy variables. Suppose that it needs to run some subprocess to model the relationship between its actions and those proxy variables. If we assume that optimizing the proxy variables is necessary to perform well on the base objective, then for a mesa-optimizer to be robustly aligned, it would also need to model the causal relationship between those proxy variables and the base objective, which might require additional subprocesses. Moreover, the benefit to the base optimizer of adding those subprocesses depends on the mesa-optimizer having additional subprocesses to model the relationship between its actions and those proxy variables. This informal argument suggests that if a mesa-optimizer's computation neatly factors in this way, then developing a robustly aligned mesa-objective may require strictly more subprocesses than developing a pseudo-aligned mesa-objective.

This suggests that, at least in a local optimization process, mesa-optimizers might tend to start their development as proxy aligned before becoming robustly aligned. In other words, rather than simultaneously gaining competence and becoming aligned, we might expect such a system to first become competent at optimizing proxies, then possibly start becoming more robustly aligned.

\textbf{Internals-based selection.} In current machine learning systems, learned algorithms are selected primarily based on how their policies perform on the base objective. There is often some selection based on the internal structure of the learned algorithm, such as weight regularization, but such selection is usually very coarse. One possible idea for addressing pseudo-alignment could be to use a selection method based on a more sophisticated function of a learned algorithm's internal workings. This approach is similar in spirit to hard-coded optimization: we use knowledge of the system's internal workings to prevent it from being proxy-aligned.

One possibility might be to develop a meta-learner that is specifically trained to prevent pseudo-alignment. While this seems beyond current meta-learning techniques, the basic principle is similar to existing meta-learners designed to optimize for robustness. An extreme form of internals-based selection might have the base optimizer reason about or prove properties of the learned algorithm as part of its selection process.

While desirable, such methods are currently out of reach primarily due to the limitations of current transparency methods. One possible concern with all of these approaches, however, is that they might require the base optimizer to be better at searching for problems than the mesa-optimizer is at hiding them. Furthermore, in order to create a base optimizer that can do this, we may need to first train it using a third optimizer, which might result in the same problem at this second layer.\footnote{Iterated Distillation and Amplification might be a possible solution to problems of this form.\cite{amplification}}

\textbf{Adversarial training.} One possible method to alleviate some of these concerns might be to sample environments adversarially from a very wide distribution. With such adversarial training, it might be possible to overcome these hurdles and make the robustly aligned objective the most reachable one. This would require making it so that, for every mesa-objective $O_\text{mesa}$ that is more reachable than the base objective $O_\text{base}$, there is some situation where $O_\text{mesa}$ and $O_\text{base}$ disagree and the mesa-optimizer is incentivized to agree with $O_\text{base}$.

If we model reachability of an objective as simply its length in bits, then distinguishing $O_\text{base}$ from every single more reachable $O_\text{mesa}$ gets exponentially harder as $O_\text{base}$ gets more complex. Thus, for a very complicated $O_\text{base}$, sufficiently incentivizing the base optimizer to find a mesa-optimizer with that $O_\text{base}$ is likely to be very difficult, though not impossible.

Even in such a case, however, there would likely still be a period of time where the learned algorithm is a misaligned mesa-optimizer, leaving open an ominous possibility: the misaligned mesa-optimizer could figure out the correct actions to take based on $O_\text{base}$ while its objective function was still $O_\text{mesa}$. We will call this situation \textit{deceptive alignment} and will discuss it at greater length in the next section.

\section{Deceptive alignment}
\label{sec:4}
\cftchapterprecistoc{One particularly concerning form of pseudo-alignment is deceptive alignment, in which the mesa-optimizer learns to model the base objective enough to optimize for it as an instrumental goal without internalizing the base objective in its mesa-objective.}

With enough training in sufficiently diverse environments, it seems plausible that the base objective will eventually have to be fully represented in the mesa-optimizer. We propose that this can happen without the mesa-optimizer becoming robustly aligned, however. Specifically, a mesa-optimizer might come to \textit{model} the base objective function and learn that the base optimizer will modify the mesa-optimizer if the mesa-optimizer scores poorly on the base objective.

If the mesa-optimizer has an objective that extends across parameter updates, then it will be incentivized to avoid being modified,\footnote{Though for clarity we speak of the same agent persisting across parameter updates, neither we nor the agent must reason in this way. A more general description of the situation is that there is a sequence of ``successor agents'' connected by parameter updates. The agent can potentially learn the effects of the parameter updates on its successors, and thus on the multi-episode return it is optimizing.} as it might not pursue the same objective after modification (with the result that its current objective will not be achieved in future iterations). This means that the mesa-optimizer will be instrumentally incentivized to act as if it is optimizing the base objective function, even if its actual mesa-objective is something else entirely. We will refer to this hypothetical phenomenon as \textit{deceptive alignment.}\footnote{The concept of deceptive alignment is closely related to Manheim and Garrabrant's concept of ``adversarial Goodhart.''\cite{goodhart} In ``Categorizing Variants of Goodhart's Law,'' Manheim and Garrabrant describe adversarial Goodhart as when ``an agent with different goals than the regulator'' causes a ``collapse of the statistical relationship between a goal which the optimizer intends and the proxy used for that goal.'' Deceptive alignment can be seen as an instance of adversarial Goodhart in which the mesa-optimizer plays the role of the agent with different goals and the base optimizer plays the role of the regulator. Training performance is the regulator's proxy, and the deployment performance is the regulator's actual goal.} Deceptive alignment is a form of instrumental proxy alignment, as fulfilling the base objective is an instrumental goal of the mesa-optimizer.

\begin{figure}[h!]
  \centering
  \includegraphics[width=\textwidth]{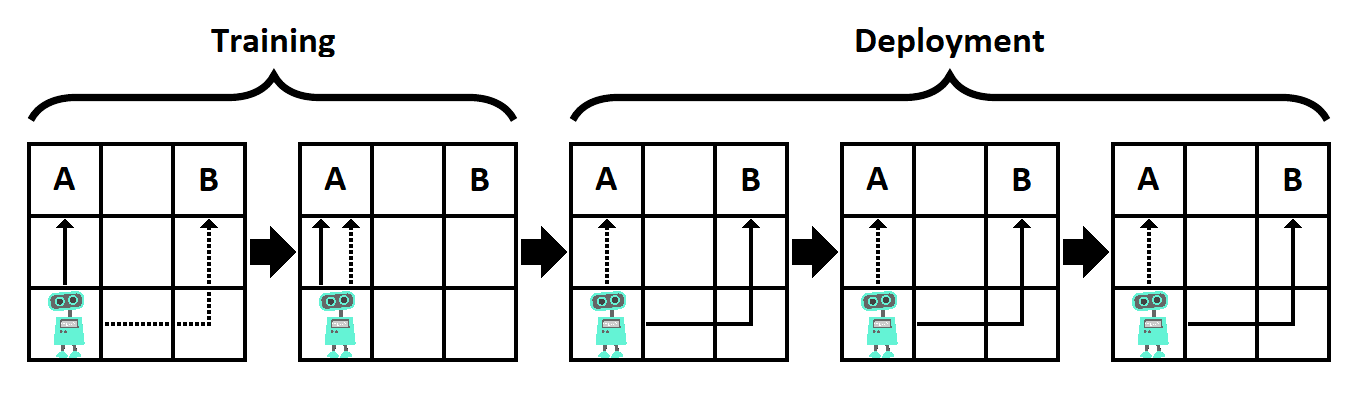}
  \caption{A toy example of deceptive alignment. In this task, the base objective is for the robot to get to A, while the mesa-objective is to get to B. The task is run for two training episodes and three deployment episodes. Parameter updates are made only after the training episodes, while after deployment the parameters are fixed. If the mesa-optimizer goes to its objective (B) during training, it will be modified to ensure it goes to A in future episodes (dashed arrows). Therefore, if the mesa-optimizer's objective spans across episodes, the dominant strategy for the mesa-optimizer is to go to A in training and B in testing (filled arrows).}
  \label{fig:4.1}
\end{figure}

As a simple example, consider an environment that consists of a $3 \times 3$ grid (see \cref{fig:4.1}). The top left corner is ``corner A'' and the top right corner is ``corner B.'' The agent, which is a mesa-optimizer, starts in the bottom left corner. The base objective is to get to corner A in as few steps as possible, but the objective of the mesa-optimizer is to get to corner B. The environment will be run five times. The agent is modified only after the first two episodes (representing the training period), but not thereafter (representing testing/deployment). If a suboptimal move is made in training, the agent is modified so that its mesa-objective becomes the base objective. The objective of the mesa-optimizer spans all episodes: it wants B to be reached in as many episodes as possible.

For the sake of this example, we will assume that the agent knows all of the above facts. If the agent goes to B in any training episode, then it will be modified to go to A in all subsequent episodes, which means that it would reach B in only one episode in total. Alternatively, if the agent goes to A in the first two episodes, then it will not be modified, and could therefore go to B in the remaining three episodes. Using this strategy, B is reached in three episodes in total, which is better (on the mesa-objective) than the one time B is reached if the mesa-optimizer goes there in training. This simple example demonstrates how a misaligned mesa-optimizer could be incentivized to ``pretend'' to be aligned with the base objective until it is no longer in training or is otherwise given an opportunity to in some way compromise or escape the training process.

We can distinguish two ways that a base objective function could be incorporated into a mesa-optimizer:\footnote{Thanks to Paul Christiano for the internalization/modeling distinction.}
\begin{enumerate}
\item \textbf{Internalization of the base objective.} The mesa-objective function gets adjusted towards the base objective function to the point where it is robustly aligned.
\item \textbf{Modeling of the base objective.} The base objective is incorporated into the mesa-optimizer's epistemic model rather than its objective, leaving open the possibility that the objective could still be misaligned.
\end{enumerate}
In the former case, information about the base objective flows into the learned algorithm via the optimization performed by the base optimizer---the base objective is built into the mesa-optimizer as it is adapted by the base optimizer. But there is another possibility: information about the base objective could enter through the mesa-optimizer's input channel instead. In this case, the base objective will be represented in the learned algorithm in a different---and potentially more unstable---way than if the base objective had been incorporated through optimization.\footnote{Note, however, that it is possible for information about the base objective to be introduced via optimization, but not form part of the mesa-objective. For example, it could form part of a prior on tasks, having no direct bearing on the mesa-objective.} In such a case, the information describing the base objective does not become internalized in the mesa-objective, potentially leading to deceptive alignment.

\subsection{Safety concerns arising from deceptive alignment}
\label{sec:4.1}
\cftchapterprecistoc{How could a deceptively aligned mesa-optimizer pose a catastrophic risk?}

If deceptive alignment is a real possibility, it presents a critical safety problem. A deceptively aligned mesa-optimizer is likely to stop optimizing the base objective as soon as it is in a position where it no longer needs to perform well on the base objective to achieve its goal. Thus, it will eventually stop performing the function it was trained to perform, and instead start pursuing some other, indeterminate goal. We will refer to this transition to optimizing for the mesa-objective instead of the base-objective as ``defection''. This is likely to happen whenever the mesa-optimizer believes defecting will no longer cause it to be modified.

Particularly concerning is the case where the mesa-optimizer actively intervenes to remove the threat of modification---by exploiting vulnerabilities in the software it is running on, for example. In extreme cases, a deceptive mesa-optimizer might try to manipulate or plan around its programmers, since they form part of the modification threat---they could shut the system down if it fails to perform well, or simply choose not to deploy it. Both of these possibilities reduce the system's expected future influence, and are therefore likely to have a low ranking in its mesa-objective. In its most extreme form, this failure of alignment could lead to a ``treacherous turn'' as described by Nick Bostrom in \textit{Superintelligence.}\cite{superintelligence}

Paul Christiano provides a scenario that illustrates how deceptive alignment could pose a catastrophic risk.\cite{paul_doom} In Christiano's example, deceptively aligned mesa-optimizers\footnote{Christiano refers to deceptively aligned mesa-optimizers as ``influence-seeking systems.''} arise from future machine learning techniques that are advanced enough to search over a large model space that includes policies that can understand the world. In this scenario, some cases of defection by deceptively aligned mesa-optimizers may have catastrophic consequences---when the systems control large-scale infrastructure, for example---though such failures will likely still be recoverable. As we rely more on automated systems to keep track of an increasingly complex world, however, it will eventually become impossible to recover from a correlated failure of many AI systems simultaneously. At this point, a cascade of defections by many deceptively aligned mesa-optimizers simultaneously could lead to the world ``going out with a bang.''\cite{paul_doom} Christiano notes that this may be most likely to first occur during a ``period of heightened vulnerability'' such as an inter-state conflict or natural disaster.

\subsection{Conditions for deceptive alignment}
\label{sec:4.2}
\cftchapterprecistoc{What conditions are necessary for a mesa-optimizer to become deceptively aligned?}

We propose that for deceptive alignment to occur the following conditions must be satisfied:
\begin{enumerate}
\item The mesa-optimizer must have an objective that extends across parameter updates.\footnote{Or must otherwise have some method of achieving more value from defecting later rather than defecting now.}
\item The mesa-optimizer must be able to model the fact that it is being selected to achieve a particular base objective and must have some model of what that objective is.\footnote{Note that it is \textit{not} required that the mesa-optimizer be able to model (or infer the existence of) the base optimizer; it only needs to model the optimization pressure it is subject to. }
\item The mesa-optimizer must expect the threat of modification\footnote{Alternatively, the threat could just be that the mesa-optimizer is never selected by the base optimizer in the first place. In that case, this condition would be trivially satisfied, as the threat to a mesa-optimizer of not being generated by the base optimizer at all would go away as soon as the base optimizer does in fact generate it.} to eventually go away, either due to training ending or because of actions taken by the mesa-optimizer.
\end{enumerate}
If the first condition is not satisfied, then the mesa-optimizer has no incentive to influence the outcome of parameter updates and will just always defect. If the second condition is not satisfied, then the mesa-optimizer will not see a reason to sacrifice gains in its current episode, as it expects future gains to be independent of what happens in this episode. If the third condition is not satisfied, then the mesa-optimizer will not expect any benefit from temporarily acting in accordance with the base objective, and will thus either always defect or never defect. One must distinguish the question of whether these conditions are possible from whether they are likely, however. We examine the factors involved in each of these conditions below.

\textbf{The mesa-optimizer must have an objective that extends across parameter updates.} It is not clear under what conditions such objectives would be incentivized. The most plausible scenario is likely one in which such an objective is encouraged by the training setup, as there seem to be situations in which this would be required for good performance. For example, the agent might be tasked with solving an environment in which its actions can have permanent consequences across episodes or where planning between episodes is otherwise necessary.

\textbf{The mesa-optimizer must be able to model the fact that it is being selected to achieve a particular base objective and must have some model of what that objective is.} A mesa-optimizer has several ways of learning about the optimization pressure it is under.\footnote{We will generally be assuming that the selection pressure takes the form of the base optimizer modifying the mesa-optimizer post-hoc, as we are mostly thinking about local optimization processes, though one could also imagine a base optimizer that instead develops a completely new inner algorithm at each time step. Though some of these issues might still apply even in such a case, we will generally not be considering such base optimizers in this section.}

First, such knowledge can come about as a result of the optimization pressure itself. Small pieces of information about the base objective can be beneficial if they enable the mesa-optimizer to improve its score on the base objective function---by improving its prior on the tasks it is likely to receive, for example. Since each piece of information independently helps with achieving the base objective, this forms an approximately monotonically increasing path that a local base optimizer might follow. The parameter updates could, in principle, be learned like any other element of the dynamics, without needing an explicit model of the base optimizer, potentially making this option available even for mesa-optimizers with quite modest capabilities.

Second, in many realistic applications of advanced machine learning, a mesa-optimizer could use information from its environment to deduce its situation at runtime. This is how humans learned that we were generated by the base optimizer of evolution, and seems like a very plausible pathway for a hypothetical mesa-optimizer to discover the existence of the optimization pressure it is under. For example, an agent tasked with forecasting stock performance in the tech sector might be given data on current AI technology in order to improve its performance, but this information might also allow it to learn about its own situation.

This suggests that it may be desirable for the programmers to limit the mesa-optimizer's access to information describing the base objective. It is unclear to what extent this is possible, given that many advanced ML systems will likely have to operate in environments where this information is readily available.

Even if the programmers successfully limit access to this information, however, there may still be other channels through which information about the base objective could enter the mesa-optimizer in unintended ways:
\begin{itemize}
\item The mesa-optimizer could reason about why it is given the particular tasks it is being asked to solve. Most mesa-optimizers would need a prior over the sorts of tasks they are likely to receive, and a mesa-optimizer might be able to learn about the existence of the base optimizer by inspecting this prior.
\item The mesa-optimizer might be able to reason about itself and about why it is the kind of thing that it is, and observe that it is well-designed to accomplish some goal.
\end{itemize}
However, these two strategies might require a much more advanced mesa-optimizer.

\textbf{The mesa-optimizer must expect the threat of modification to eventually go away, either due to training ending or because of actions taken by the mesa-optimizer.} First, note that this condition is a statement about the mesa-optimizer's expectations, not necessarily the actual truth. Second, even if there is no point at which training stops by design, so long as there exists some way for the mesa-optimizer to execute a treacherous turn, there is always some way for it to remove the threat of modification. For the mesa-optimizer to act on such a strategy, however, it has to actually be aware of some way for it to do so. Conversely, a mesa-optimizer that mistakenly believes it has an opportunity to defect might also exhibit deceptive alignment.

Since it seems difficult to discover through exploration that the threat of modification can disappear, the mesa-optimizer will either have to infer this through some form of abstract reasoning or have a prior that assigns a non-negligible probability to the hypothesis that the selection pressure will not continue indefinitely. The second option seems plausible; a Bayesian prior should always assign a nonzero probability to the hypothesis that some fact will eventually stop being true.\footnote{Note that if the mesa-optimizer also believes that the opportunities it currently has for achieving its own objective are unusually good then it might still pursue its objective greedily, but this would presumably not always be the case.} As for the first option, the mesa-optimizer could infer that the optimization pressure at some point may disappear if it learns about the existence of the base optimizer (either by using information in its input or by having a strong prior for the existence of a base optimizer), though this might require a relatively advanced mesa-optimizer.

\subsection{The learning dynamics of deceptive alignment}
\label{sec:4.3}
\cftchapterprecistoc{How would a deceptively aligned mesa-optimizer behave during the training process?}

Once a mesa-optimizer becomes deceptive, the dynamics of what causes it to perform well on the base objective begin to work differently. Instead of being selected only on the basis of its objective and ability, a deceptively aligned optimizer is also selected for how well it infers the base objective function and how discerning it is about when to optimize the base objective versus its own mesa-objective. We identify at least two regimes of selection for a mesa-optimizer under deceptive alignment.

\textbf{Joint optimization.} If the mesa-optimizer attempts to jointly optimize its objective and the base objective or makes mistakes about the timing of its defection, then there will still be some pressure on its objective to be more compatible with the base objective. This is unlikely to be the same pressure as exists when the optimizer is just working towards its own objective, but there might be some movement towards an objective that is more compatible with being jointly optimized alongside the base objective.

\textbf{Pure deception.} If the mesa-optimizer stops trying to optimize its own objective in the short term and focuses on cooperating with the selection process entirely then this may result in its objective ``crystallizing.'' As its objective is largely irrelevant to its outputs now, there is little selection pressure acting to modify this objective. As a result, the objective becomes effectively locked in, excepting random drift and any implicit time complexity or description length penalties.

The joint optimization mode is not stable because it is suboptimal from the perspective of the base optimizer. As the mesa-optimizer becomes more competent, it will therefore have to either move to the domain of pure deception or become robustly aligned.

\subsection{Internalization or deception after extensive training}
\label{sec:4.4}
\cftchapterprecistoc{Will a mesa-optimizer with extensive training still be deceptive?}

In the limit of training on a diverse set of tasks, we expect joint optimization of both the base and mesa- objectives to be unstable. Assuming that the mesa-optimizer converges towards behavior that is optimal from the perspective of the base optimizer, the mesa-optimizer must somehow learn the base objective. We identify three distinct ways in which this can happen:
\begin{enumerate}
\item \textbf{Robust alignment through internalization.} The mesa-objective function gets adjusted towards the base objective function, removing any pseudo-alignment.
\item \textbf{Crystallization of deceptive alignment.} Information about the base objective is increasingly incorporated into the mesa-optimizer's epistemic model without its objective becoming robustly aligned. The mesa-optimizer ends up fully optimizing for the base objective, but only for instrumental reasons, without its mesa-objective getting changed.
\item \textbf{Robust alignment through corrigibility.} Information about the base objective is incorporated into the mesa-optimizer's epistemic model and its objective is modified to ``point to'' that information. This situation would correspond to a mesa-optimizer that is corrigible\cite{corrigibility} with respect to the base objective (though not necessarily the programmer's intentions).
\end{enumerate}
To distinguish between the two different paths to robust alignment, we will use the term \textit{internally aligned} to refer to the first case and the term \textit{corrigibly aligned} to refer to the last case. We analyze some ways in which these different paths may be more or less attractive below.

\textbf{There are more paths to deceptive alignment than to robust alignment.} Since the future value of its objective depends on the parameter updates, a mesa-optimizer that meets the three criteria for deceptive alignment is likely to have a strong incentive to understand the base objective better. Even a robustly aligned mesa-optimizer that meets the criteria is incentivized to figure out the base objective in order to determine whether or not it will be modified, since before doing so it has no way of knowing its own level of alignment with the base optimizer. Mesa-optimizers that are capable of reasoning about their incentives will, therefore, attempt to get more information about the base objective. Furthermore, once a mesa-optimizer learns about the base objective, the selection pressure acting on its objective will significantly decrease, potentially leading to a crystallization of the mesa-objective. However, due to unidentifiability (as discussed in \cref{sec:3}), most mesa-objectives that are aligned on the training data will be pseudo-aligned rather than robustly aligned. Thus, the most likely sort of objective to become crystallized is a pseudo-aligned one, leading to deceptive alignment.

\textbf{Corrigible alignment seems to require already having a model of the base objective.} For corrigible alignment to be beneficial from the perspective of the base optimizer, the mesa-optimizer has to already have some model of the base objective to ``point to.'' However, once a mesa-optimizer has a model of the base objective, it is likely to become deceptively aligned---at least as long as it also meets the other conditions for deceptive alignment. Once a mesa-optimizer becomes deceptive, it will remove most of the incentive for corrigible alignment, however, as deceptively aligned optimizers will also behave corrigibly with respect to the base objective, albeit only for instrumental reasons.

\textbf{Deceptive and corrigible alignment might be more computationally expensive.} In the case of deceptive alignment, simultaneously working on optimizing the base objective and working on identifying good opportunities to defect is likely to require more computation than just doing the former. In a case where the total amount of computation available is fixed, a deceptively aligned mesa-optimizer is going to have less time to spend on optimizing the base objective than any robustly aligned optimizer, resulting in potentially worse plans with respect to the base objective. Furthermore, in both deceptive and corrigible alignment, the mesa-optimizer will have to spend time learning about the base objective to enable it to properly optimize for it, making internal alignment the most time efficient. Thus, an explicit or implicit time penalty would seem to disadvantage a deceptively aligned mesa-optimizer and encourage an internally aligned one. However, it is unclear what the magnitude of this extra cost is. It is plausible that on complex tasks, the overhead of spotting defection opportunities could be low in comparison to the cost of solving the task itself. Moreover, even if it is not globally optimal, incurring this overhead might be a local optimum, with no clear paths for the base optimizer to escape it.

\textbf{Deceptive and corrigible alignment might have a lower description length.} Deceptively and corrigibly aligned mesa-optimizers are likely to have an advantage in representing the base objective compared to internally aligned mesa-optimizers as they are able to use the environment to provide some of the information. Specifically, a deceptively or corrigibly aligned mesa-optimizer doesn't need to represent the entire base objective internally, as it can infer pieces of it by looking for information in its environment during each run. An internally aligned mesa-optimizer, by contrast, has to start with a full model of the base objective in its mesa-objective to make it robustly aligned, since an internally aligned mesa-optimizer will have no incentive to change its objective based on any additional information it gets about the base objective. Thus, to the extent that a base optimizer is selecting based on description length, it might preferentially select for a deceptively or corrigibly aligned mesa-optimizer over an internally aligned one.

\textbf{Modeling might happen faster than internalization.} It may take less time for the mesa-optimizer to learn to model the base objective than for the base optimizer to adapt the mesa-objective. Suppose that a mesa-optimizer is deceptive but has not yet correctly inferred the full base objective.\footnote{An analogous argument applies for a deceptively aligned mesa-optimizer that is defecting too early, or one that unsuccessfully attempts a treacherous turn.} In such a case, what will determine whether it ends up robustly aligned or deceptively aligned after further training will be whether it infers enough about the base objective before the base optimizer modifies its mesa-objective (either to make it internally aligned or corrigibly aligned). Two factors are likely to contribute to which happens first: the rate at which the mesa-optimizer improves its estimate of the base objective, and the parameter updates made by the base optimizer at each mistake.

The first factor, improving its estimate of the base objective, seems to depend primarily on the capabilities of the mesa-optimizer during the timeframe when it is deceptive, and on the complexity of the base objective that needs to be inferred. An important special case is when the mesa-optimizer has access to cross-episodic state. Such state could be used as memory, allowing the mesa-optimizer to build upon its prior thinking in a process that happens faster than the rate at which the parameter updates align its mesa-objective.

The second factor, parameter updates, depends on the nature of the base optimizer. Current gradient-descent-based base optimizers would have a small reaction to any given mistake, but more sophisticated base optimizers with advanced safety features might be able to do better. For example, a more sophisticated base optimizer might be able to scramble the mesa-optimizer's parameters or call for human oversight whenever a suspicious failure is detected.

In general, parameter updates are not guaranteed to make a mesa-optimizer more aligned. For instance, instead of aligning the mesa-objective, an update made after a defection could push the system to improve its planning to the point that it no longer makes detectable mistakes.

A consequence of these considerations is that any possible systematic bias in how the base optimizer updates the mesa-optimizer's objective or modeling capabilities could play a role in determining the final outcome.\footnote{Furthermore, a deceptively aligned mesa-optimizer would be incentivized to cause there to be a systematic bias in the direction of preventing the base optimizer from modifying its mesa-objective. Thus, in the context of a local optimization process, a deceptive mesa-optimizer might try to ``hack'' its own gradient (by, for example, making itself more brittle in the case where its objective gets changed) to ensure that the base optimizer adjusts it in such a way that leaves its mesa-objective untouched.} This analysis remains speculative, however---if we are to have confidence in systems that are capable of producing mesa-optimizers, we still need a better understanding of the conditions under which strongly optimizing a system results in deceptive alignment rather than robust alignment.

\subsection{Distributional shift and deceptive alignment}
\label{sec:4.5}
\cftchapterprecistoc{What happens when a deceptively aligned mesa-optimizer undergoes distributional shift?}

Once a mesa-optimizer is deceptive, it will begin to monitor its input for opportunities to defect. During training, a deceptively aligned optimizer is strongly discouraged from defecting because it knows that the base optimizer will select against it. However, once a deceptively aligned mesa-optimizer is free of the threat of external modification, this no longer applies. Thus, a deceptively aligned mesa-optimizer will be looking for ways to determine whether or not it is in training---that is, for the presence of a distributional shift.

This means that the detrimental effects of deceptive alignment will only be externally visible when the mesa-optimizer defects---either randomly or after undergoing a distributional shift. However, the underlying problem is not the distributional shift itself, but the failure to align the mesa-optimizer in the first place. One can either try to solve the underlying problem, by making the mesa-optimizer robust to the distributional shift (i.e., robustly aligned), or one can try to patch the problem by preventing any distributional shift from occurring.

In current AI systems, a small amount of distributional shift between training and deployment need not be problematic: so long as the difference is small enough in the task-relevant areas, the training distribution does not need to perfectly reflect the deployment distribution. However, this may not be the case for a deceptively aligned mesa-optimizer. If a deceptively aligned mesa-optimizer is sufficiently advanced, it may detect very subtle distributional shifts for the purpose of inferring when the threat of modification has ceased.

This raises the question of whether it would even be possible to contain a deceptive mesa-optimizer by completely removing any detectable difference between training and deployment. There are a number of reasons to believe that this might not be feasible in practice.

First, it would significantly limit the use cases for the system. Machine learning is often done with the intent of making systems that can solve new problems. If we require that there is no distributional shift of any form, then we cannot allow the system to encounter problems that it could not have encountered during training, which means that we cannot give it truly new problems. For example, training the system in a simulated environment and deploying it in the real world would likely involve enough detectable distributional shift to make it infeasible under this approach.

Second, there are subtle forms of distributional shift that seem very difficult to eliminate. For example, if the mesa-optimizer is able to detect the presence of parameter updates---via any state carried over between steps, for instance---then it would always experience a form of distributional shift at deployment. Running the system for a longer amount of time (or giving the system access to some form of long-term memories) would also constitute a form of distributional shift. Some examples of differences that a mesa-optimizer might be able to detect include:
\begin{itemize}
\item Training in a simulated environment and deploying in the real world.
\item Running the system for a different amount of time during training and testing.
\item If a model is able to learn from its past experiences, then having access to more memories, or having access to the kinds of memories it would form at later time steps (if there is a difference between the two).
\item The presence of adversarial-example-type attacks by an adversary during deployment.
\item Encountering something that is very hard to compute during training, but that is likely to eventually be computed during deployment (e.g. a factorization of RSA 2048\cite{paul_robust_corrigibility}).
\item The ability of the base optimizer to modify the learned algorithm during training but not at other times.
\item The ability of the programmers to correct for perceived errors during training and testing but not deployment.
\item The presence or absence of good opportunities for the mesa-optimizer to defect against its programmers.
\end{itemize}
Furthermore, it should be noted that it is not strictly necessary that a mesa-optimizer be able to detect the difference between training and deployment in order for it to be incentivized to behave deceptively. Consider again the example in \cref{fig:4.1}, but suppose this time that the mesa-optimizer is unable to detect which episode it is currently in. The mesa-optimizer is then facing a decision problem similar to the ``absent-minded driver problem.''\cite{absent_minded_driver} If we apply the same reasoning as in the absent-minded driver problem to our example in \cref{fig:4.1}, the planning-optimal policy for the mesa-optimizer is to defect with probability $p \approx 0.61$. Thus, the mesa-optimizer need not be able to detect when training has ended for it to be incentivized to ``gamble'' and randomly defect with some probability. Furthermore, a deceptively aligned mesa-optimizer might also try to look for opportunities to defect that would be exceptionally good conditional on it currently not being trained, but that are rare enough to be unlikely to come up during training.

\section{Related work}
\label{sec:5}
\cftchapterprecistoc{How does mesa-optimization relate to other fields such as meta-learning and reward learning?}

\textbf{Meta-learning.} As described in \cref{sec:1}, meta-learning can often be thought of meta-optimization when the meta-optimizer's objective is explicitly designed to accomplish some base objective. However, it is also possible to do meta-learning by attempting to make use of mesa-optimization instead. For example, in Wang et al.'s ``Learning to Reinforcement Learn,'' the authors claim to have produced a neural network that implements its own optimization procedure.\cite{learn_to_rl} Specifically, the authors argue that the ability of their network to solve extremely varied environments without explicit retraining for each one means that their network must be implementing its own internal learning procedure. Another example is Duan et al.'s ``$RL^2$: Fast Reinforcement Learning via Slow Reinforcement Learning,'' in which the authors train a reinforcement learning algorithm which they claim is itself doing reinforcement learning.\cite{rl2} This sort of meta-learning research seems the closest to producing mesa-optimizers of any existing machine learning research.

\textbf{Robustness.} A system is robust to distributional shift if it continues to perform well on the objective function for which it was optimized even when off the training environment.\cite{concrete_problems} In the context of mesa-optimization, pseudo-alignment is a particular way in which a learned system can fail to be robust to distributional shift: in a new environment, a pseudo-aligned mesa-optimizer might still competently optimize for the mesa-objective but fail to be robust due to the difference between the base and mesa- objectives.

The particular type of robustness problem that mesa-optimization falls into is the reward-result gap, the gap between the reward for which the system was trained (the base objective) and the reward that can be reconstructed from it using inverse reinforcement learning (the behavioral objective).\cite{leike} In the context of mesa-optimization, pseudo-alignment leads to a reward-result gap because the system's behavior outside the training environment is determined by its mesa-objective, which in the case of pseudo-alignment is not aligned with the base objective.

It should be noted, however, that while inner alignment is a robustness problem, the occurrence of unintended mesa-optimization is not. If the base optimizer's objective is not a perfect measure of the human's goals, then preventing mesa-optimizers from arising at all might be the preferred outcome. In such a case, it might be desirable to create a system that is strongly optimized for the base objective within some limited domain without that system engaging in open-ended optimization in new environments.\cite{drexler} One possible way to accomplish this might be to use strong optimization at the level of the base optimizer during training to prevent strong optimization at the level of the mesa-optimizer.\cite{drexler}

\textbf{Unidentifiability and goal ambiguity.} As we noted in \cref{sec:3}, the problem of unidentifiability of objective functions in mesa-optimization is similar to the problem of unidentifiability in reward learning, the key issue being that it can be difficult to determine the ``correct'' objective function given only a sample of that objective's output on some training data.\cite{irl_unidentifiability} We hypothesize that if the problem of unidentifiability can be resolved in the context of mesa-optimization, it will likely (at least to some extent) be through solutions that are similar to those of the unidentifiability problem in reward learning. An example of research that may be applicable to mesa-optimization in this way is Amin and Singh's\cite{irl_unidentifiability} proposal for alleviating empirical unidentifiability in inverse reinforcement learning by adaptively sampling from a range of environments.

Furthermore, it has been noted in the inverse reinforcement learning literature that the reward function of an agent generally cannot be uniquely deduced from its behavior.\cite{armstrong_preferences} In this context, the inner alignment problem can be seen as an extension of the value learning problem. In the value learning problem, the problem is to have enough information about an agent's behavior to infer its utility function, whereas in the inner alignment problem, the problem is to test the learned algorithm's behavior enough to ensure that it has a certain objective function.

\textbf{Interpretability.} The field of interpretability attempts to develop methods for making deep learning models more interpretable by humans. In the context of mesa-optimization, it would be beneficial to have a method for determining whether a system is performing some kind of optimization, what it is optimizing for, and/or what information it takes into account in that optimization. This would help us understand when a system might exhibit unintended behavior, as well as help us construct learning algorithms that create selection pressure against the development of potentially dangerous learned algorithms.

\textbf{Verification.} The field of verification in machine learning attempts to develop algorithms that formally verify whether systems satisfy certain properties. In the context of mesa-optimization, it would be desirable to be able to check whether a learned algorithm is implementing potentially dangerous optimization.

Current verification algorithms are primarily used to verify properties defined on input-output relations, such as checking invariants of the output with respect to user-definable transformations of the inputs. A primary motivation for much of this research is the failure of robustness against adversarial examples in image recognition tasks. There are both white-box algorithms,\cite{safety_verification} e.g. an SMT solver that in principle allows for verification of arbitrary propositions about activations in the network,\cite{reluplex} and black-box algorithms\cite{practical_verification}. Applying such research to mesa-optimization, however, is hampered by the fact that we currently don't have a formal specification of optimization.

\textbf{Corrigibility.} An AI system is \textit{corrigible} if it tolerates or assists with its human programmers in correcting itself.\cite{corrigibility} The current analysis of corrigibility has focused on how to define a utility function such that, if optimized by a rational agent, that agent would be corrigible. Our analysis suggests that even if such a corrigible objective function could be specified or learned, it is nontrivial to ensure that a system trained on that objective function would actually be corrigible. Even if the base objective function would be corrigible if optimized directly, the system may exhibit mesa-optimization, in which case the system's mesa-objective might not inherit the corrigibility of the base objective. This is somewhat analogous to the problem of utility-indifferent agents creating other agents that are not utility-indifferent.\cite{corrigibility} In \cref{sec:4}, we suggest a notion related to corrigibility---corrigible alignment---which is applicable to mesa-optimizers. If work on corrigibility were able to find a way to reliably produce corrigibly aligned mesa-optimizers, it could significantly contribute to solving the inner alignment problem.

\textbf{Comprehensive AI Services (CAIS).}\cite{drexler} CAIS is a descriptive model of the process by which superintelligent systems will be developed, together with prescriptive implications for the best mode of doing so. The CAIS model, consistent with our analysis, makes a clear distinction between learning (the base optimizer) and functionality (the learned algorithm). The CAIS model predicts, among other things, that more and more powerful general-purpose learners will be developed, which through a layered process will develop services with superintelligent capabilities. Services will develop services that will develop services, and so on. At the end of this ``tree,'' services for a specific final task are developed. Humans are involved throughout the various layers of this process so that they can have many points of leverage for developing the final service.

The higher-level services in this tree can be seen as meta-optimizers of the lower-level services. However, there is still the possibility of mesa-optimization---in particular, we identify two ways in which mesa-optimization could occur in the CAIS-model. First, a final service could develop a mesa-optimizer. This scenario would correspond closely to the examples we have discussed in this paper: the base optimizer would be the next-to-final service in the chain, and the learned algorithm (the mesa-optimizer in this case), would be the final service (alternatively, we could also think of the entire chain from the first service to the next-to-final service as the base optimizer). Second, however, an intermediary service in the chain might also be a mesa-optimizer. In this case, this service would be an optimizer in \textit{two respects:} it would be the meta-optimizer of the service below it (as it is by default in the CAIS model), but it would also be a mesa-optimizer with respect to the service above it.

\section{Conclusion}
\cftchapterprecistoc{More research is needed to understand the nature of mesa-optimization in order to properly address mesa-optimization-related safety concerns in advanced ML systems.}

In this paper, we have argued for the existence of two basic AI safety problems: the problem that mesa-optimizers may arise even when not desired (unintended mesa-optimization), and the problem that mesa-optimizers may not be aligned with the original system's objective (the inner alignment problem). However, our work is still only speculative. We are thus left with several possibilities:
\begin{enumerate}
\item If mesa-optimizers are very unlikely to occur in advanced ML systems (and we do not develop them on purpose), then mesa-optimization and inner alignment are not concerns.
\item If mesa-optimizers are not only likely to occur but also difficult to prevent, then solving both inner alignment and outer alignment becomes critical for achieving confidence in highly capable AI systems.
\item If mesa-optimizers are likely to occur in future AI systems by default, and there turns out to be some way of preventing mesa-optimizers from arising, then instead of solving the inner alignment problem, it may be better to design systems to not produce a mesa-optimizer at all. Furthermore, in such a scenario, some parts of the outer alignment problem may not need to be solved either: if an AI system can be prevented from implementing any sort of optimization algorithm, then there may be more situations where it is safe for the system to be trained on an objective that is not perfectly aligned with the programmer's intentions. That is, if a learned algorithm is not an optimizer, it might not optimize the objective to such an extreme that it would cease to produce positive outcomes.
\end{enumerate}

Our uncertainty on this matter is a potentially significant hurdle to determining the best approaches to AI safety. If we do not know the relative difficulties of the inner alignment problem and the unintended optimization problem, then it is unclear how to adequately assess approaches that rely on solving one or both of these problems (such as Iterated Distillation and Amplification\cite{amplification} or AI safety via debate\cite{debate}). We therefore suggest that it is both an important and timely task for future AI safety work to pin down the conditions under which the inner alignment problem and the unintended optimization problem are likely to occur as well as the techniques needed to solve them.

\section{Glossary}

\subsection{Section 1 Glossary}
\begin{itemize}
\item \textbf{Base optimizer:} A \textit{base optimizer} is an optimizer that searches through algorithms according to some objective.
  \begin{itemize}
  \item \textbf{Base objective:} A \textit{base objective} is the objective of a base optimizer.
  \end{itemize}
\item \textbf{Behavioral objective:} The \textit{behavioral objective} is what an optimizer appears to be optimizing for. Formally, the behavioral objective is the objective recovered from perfect inverse reinforcement learning.
\item \textbf{Inner alignment:} The \textit{inner alignment problem} is the problem of aligning the base and mesa- objectives of an advanced ML system.
\item \textbf{Learned algorithm:} The algorithms that a base optimizer is searching through are called \textit{learned algorithms.}
\item \textbf{Mesa-optimizer:} A \textit{mesa-optimizer} is a learned algorithm that is itself an optimizer.
  \begin{itemize}
  \item \textbf{Mesa-objective:} A \textit{mesa-objective} is the objective of a mesa-optimizer.
  \item \textbf{Mesa-optimization:} \textit{Mesa-optimization} refers to the situation in which a learned algorithm found by a base optimizer is itself an optimizer.
  \end{itemize}
\item \textbf{Meta-optimizer:} A \textit{meta-optimizer} is a system which is tasked with producing a base optimizer.
\item \textbf{Optimizer:} An \textit{optimizer} is a system that internally searches through some space of possible outputs, policies, plans, strategies, etc. looking for those that do well according to some internally-represented objective function.
\item \textbf{Outer alignment:} The \textit{outer alignment problem} is the problem of aligning the base objective of an advanced ML system with the desired goal of the programmers.
\item \textbf{Pseudo-alignment:} A mesa-optimizer is \textit{pseudo-aligned} with the base objective if it appears aligned on the training data but is not robustly aligned.
\item \textbf{Robust alignment:} A mesa-optimizer is \textit{robustly aligned} with the base objective if it robustly optimizes for the base objective across distributions.
\end{itemize}

\subsection{Section 2 Glossary}
\begin{itemize}
\item \textbf{Local optimization process:} A \textit{local optimization process} is an optimizer that uses local hill-climbing as its means of search.
\item \textbf{Reachability:} The \textit{reachability} of a learned algorithm refers to the difficulty for the base optimizer to find that learned algorithm.
\end{itemize}

\subsection{Section 3 Glossary}
\begin{itemize}
\item \textbf{Approximate alignment:} An \textit{approximately aligned} mesa-optimizer is a pseudo-aligned mesa-optimizer where the base and mesa- objectives are approximately the same up to some degree of approximation error due to the difficulty of representing the base objective in the mesa-optimizer.
\item \textbf{Proxy alignment:} A \textit{proxy aligned} mesa-optimizer is a pseudo-aligned mesa-optimizer that has learned to optimize for some proxy of the base objective instead of the base objective itself.
  \begin{itemize}
  \item \textbf{Instrumental alignment:} \textit{Instrumental alignment} is a type of proxy alignment in which the mesa-optimizer optimizes the proxy as an instrumental goal of increasing the mesa-objective in the training distribution.
  \item \textbf{Side-effect alignment:} \textit{Side-effect alignment} is a type of proxy alignment in which optimizing for the mesa-objective has the direct causal result of increasing the base objective in the training distribution.
  \end{itemize}
\item \textbf{Suboptimality alignment:} A \textit{suboptimality aligned} mesa-optimizer is a pseudo-aligned mesa-optimizer in which some deficiency, error, or limitation causes it to exhibit aligned behavior.
\end{itemize}

\subsection{Section 4 Glossary}
\begin{itemize}
\item \textbf{Corrigible alignment:} A \textit{corrigibly aligned} mesa-optimizer is a robustly aligned mesa-optimizer that has a mesa-objective that ``points to'' its epistemic model of the base objective.
\item \textbf{Deceptive alignment:} A \textit{deceptively aligned} mesa-optimizer is a pseudo-aligned mesa-optimizer that has enough information about the base objective to seem more fit from the perspective of the base optimizer than it actually is.
\item \textbf{Internal alignment:} An \textit{internally aligned} mesa-optimizer is a robustly aligned mesa-optimizer that has internalized the base objective in its mesa-objective.
\end{itemize}

\bibliography{mesa_optimization}

\end{document}